\documentclass[1p]{elsarticle}
\usepackage[utf8]{inputenc}

\usepackage{amsmath}
\usepackage{amsfonts}
\usepackage{amssymb}

\usepackage{graphicx}
\usepackage{xcolor} 
\usepackage{svg}   
\usepackage{subcaption} 

\usepackage{booktabs} 
\usepackage{array}   
\usepackage{multirow} 
\usepackage{makecell}
\usepackage{adjustbox}
\usepackage{xcolor}
\usepackage{url} 
\usepackage{dblfloatfix} 
\usepackage{hyperref}
\usepackage{svg}
\usepackage{tabularx}

\usepackage{geometry}
\hypersetup{
    colorlinks=true,
    linkcolor=blue,
    filecolor=magenta,
    urlcolor=cyan,
    citecolor=blue, 
}

\definecolor{azul}{rgb}{0.0, 0.53, 0.74}

\journal{Computers and Electronics in Agriculture}

\begin{document}
	
\begin{frontmatter}
	
\title{SugarcaneShuffleNet: A Very Fast, Lightweight Convolutional Neural Network for Diagnosis of 15 Sugarcane Leaf Diseases}

\author[inst_rme]{Shifat E. Arman}

\author[inst_gau]{Hasan Muhammad Abdullah}
\ead{hasan.abdullah@gau.edu.bd}
\cortext[cor1]{Corresponding author}

\author[inst_rme]{Syed Nazmus Sakib}

\author[inst_bsri]{RM Saiem}

\author[inst_bsri]{Shamima Nasrin Asha}

\author[inst_rme]{Md Mehedi Hasan}

\author[inst_cse]{Shahrear Bin Amin}

\author[inst_rme]{S M Mahin Abrar}

\address[inst_rme]{Department of Robotics and Mechatronics Engineering, University of Dhaka, Dhaka 1000, Bangladesh}
\address[inst_gau]{GIS and Remote Sensing Lab, Gazipur Agricultural University, Gazipur 1706, Bangladesh}
\address[inst_bsri]{Bangladesh Sugarcrop Research Institute, Ishwardi, Pabna, Bangladesh}
\address[inst_cse]{Department of Computer Science and Engineering, University of Dhaka, Dhaka 1000, Bangladesh}

\begin{abstract}
    
Despite progress in AI-based plant diagnostics, sugarcane farmers in low-resource regions remain vulnerable to leaf diseases due to the lack of scalable, efficient, and interpretable tools. Many deep learning models fail to generalize under real-world conditions and require substantial computational resources, limiting their use in resource-constrained regions. In this paper, we present SugarcaneLD-BD, a curated dataset for sugarcane leaf-disease classification; SugarcaneShuffleNet, an optimized lightweight model for rapid on-device diagnosis; and SugarcaneAI, a Progressive Web Application for field deployment. \textbf{SugarcaneLD-BD} dataset contains 638 carefully curated images from five classes, including four major sugarcane diseases, collected in Bangladesh under diverse field conditions and verified by expert pathologists. To enhance diversity, we combined SugarcaneLD-BD with two additional datasets, yielding a larger and more representative corpus for sugarcane leaf-disease classification. Our optimized model, \textbf{SugarcaneShuffleNet}, offers the best trade-off between speed and accuracy for real-time, on-device diagnosis. This 9.26 MB model achieved 98.02\% accuracy, an F1-score of 0.98, and an average inference time of 4.14ms per image. For comparison, we fine-tuned five other lightweight convolutional neural networks: MnasNet, EdgeNeXt, EfficientNet-Lite, MobileNet, and SqueezeNet using transfer learning and Bayesian optimization. MnasNet and EdgeNeXt achieved comparable accuracy to SugarcaneShuffleNet, but required significantly more parameters, memory, and computation, limiting their suitability for low-resource deployment. Finally, we integrate SugarcaneShuffleNet into \textbf{SugarcaneAI}, a Progressive Web Application that provides farmers with Grad-CAM-based explainable diagnoses in field conditions. Together, these contributions offer a diverse benchmark, efficient models for low-resource environments, and a practical tool for sugarcane disease classification.

\end{abstract}

\end{frontmatter}
	
\section{Introduction}

Sugarcane plays a vital role in the global agricultural economy. It supports millions of smallholder farmers and supplies most of the world’s sugar and bioethanol \cite{dufey2008impacts, heinrichs2017importance}. However, foliar diseases like rust, mosaic, and brown spot can severely reduce yields. These diseases damage leaves and limit photosynthesis, which negatively influences both cane yield and sugar content \cite{huang2018diagnosis}. Even moderate infections can have a noticeable impact on crop yield and sugar quality. For this reason, early detection and intervention are key to reducing losses in the field.

Farmers and field workers usually identify sugarcane diseases by looking for symptoms like yellow bands or dark spots on leaves, followed by laboratory tests. This process takes a lot of time and effort, often making it difficult to use in rural areas with limited resources \cite{srinivasan2025sugarcane}. It also needs trained experts and can take more than a week to get results, during which the disease can spread across large areas of crops.

Convolutional neural networks (CNNs) \cite{lecun1995convolutional} offer a faster, more scalable alternative to manual inspection by automatically learning patterns such as texture and color from images, even under varied lighting and background conditions \cite{lu2021review}. Researchers have successfully applied CNNs \cite{lecun1995convolutional} and transformer-based models \cite{vaswani2017attention} to detect leaf diseases in crops such as tomato \cite{karthik2020attention}, banana \cite{bhuiyan2023bananasqueezenet}, mango \cite{hossain2024deep}, and rice \cite{yang2024novel}. These models have become reliable tools for plant disease detection and often report classification accuracies above 95\% on benchmark datasets.

Despite progress in plant disease detection, there remains a lack of publicly available datasets specific to sugarcane. Widely used datasets like PlantVillage \cite{hughes2015open} and PlantDoc \cite{singh2020plantdoc} focus mainly on vegetable and pulse crops and do not include sugarcane leaf diseases. Even existing field-based datasets are limited in class coverage, geographic diversity, or labeling consistency. A few proprietary sugarcane datasets exist, but these are often small in scale and lack representation from multiple agro-ecological zones. This scarcity of high-quality, field-representative sugarcane data makes it difficult to develop models that generalize well under real farming conditions.

Another issue with existing studies on sugarcane is the lack of lightweight models suitable for deployment in the field. Deep CNNs, while accurate, require large memory and often depend on GPU acceleration. Transformer models also demand substantial computational resources, making them unsuitable for low-power devices. This limits their use for real-time, on-device diagnosis in rural settings. There is a clear need for lightweight, efficient models that can offer high accuracy without high computational cost.

\begin{figure*}[!htpb]
    \centering
    \includegraphics[width=\textwidth]{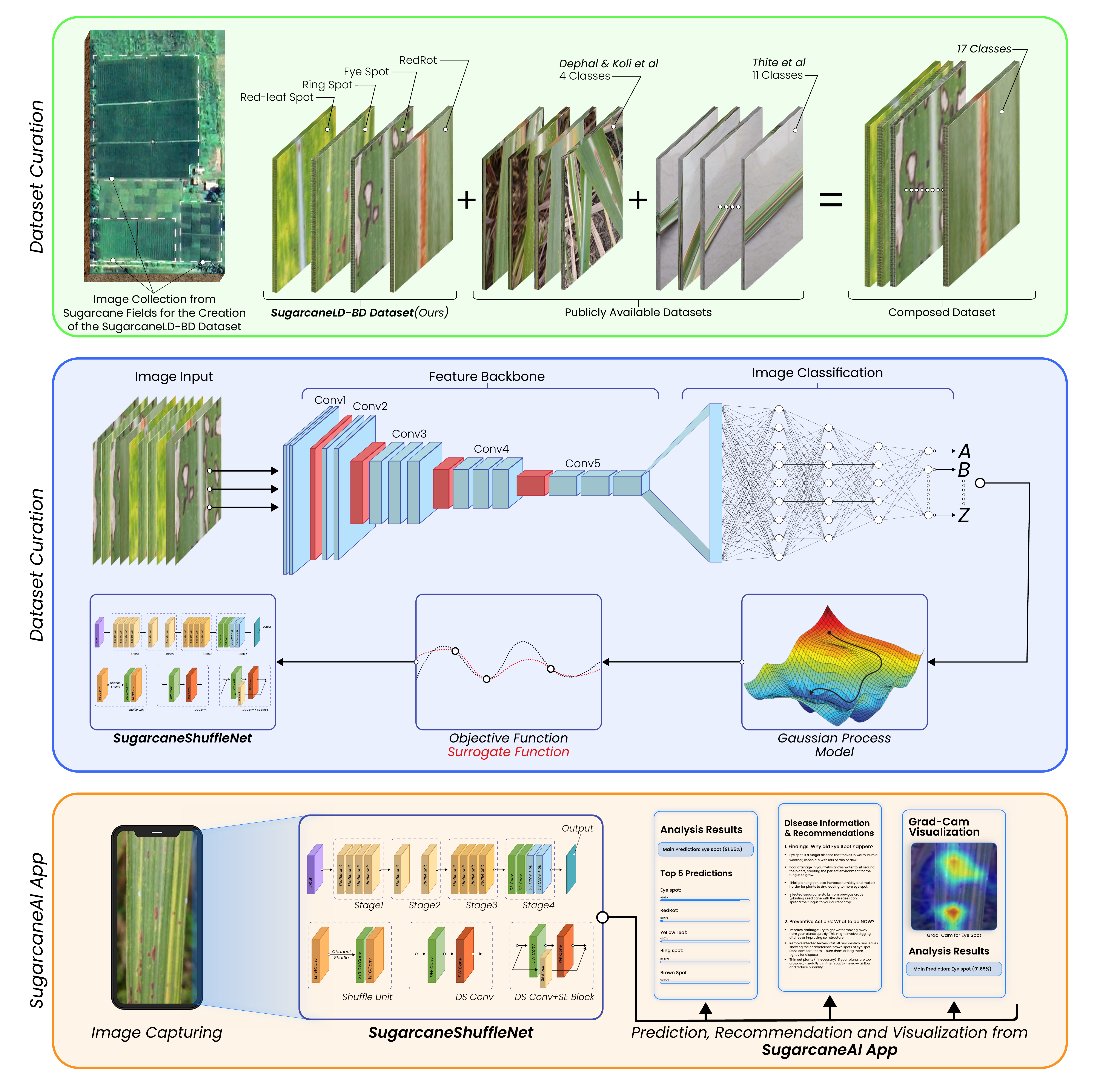}
    \caption{Overview of our sugarcane disease classification pipeline. (1) We introduce SugarcaneLD-BD, a field-representative dataset comprising 638 labeled images across four major disease classes, later combined with public datasets into a unified 17-class corpus. (2) We develop SugarcaneShuffleNet, a compact 9.26 MB model achieving 98.02\% accuracy and 4.14 ms inference latency for real-time, low-resource deployment. (3) We deploy the model in SugarcaneAI, a smartphone-compatible application providing instant disease prediction, Grad-CAM-based visual explanations, and LLM-powered agronomic guidance.}
    \label{fig:overall}
\end{figure*}

To address these challenges, we present three main contributions:

\begin{enumerate}
\item \textbf{SugarcaneLD-BD Dataset: A Field Representative Dataset for Sugarcane Leaf Disease Detection} \\
We created a carefully collected dataset of 638 images covering four major sugarcane leaf disease classes found in Bangladesh and South Asia \cite{arman2025sugarcaneld}. By combining it with two publicly available datasets, we built a larger collection of 9,908 images from 17 classes. This combined dataset includes a wide range of field backgrounds, lighting conditions, and disease stages, helping to fill an important gap in publicly available sugarcane leaf image resources.

\item \textbf{SugarcaneShuffleNet: A Rapid, Lightweight CNN for Sugarcane Leaf Disease Diagnosis} \\
We developed SugarcaneShuffleNet, a lightweight 9.26 MB model for sugarcane leaf disease classification. It achieved 98.02\% accuracy and a macro-$F_1$ score of 0.98 on a 17-class test set, with an inference time of 4.14 ms per image, enabling accurate real-time diagnosis on low-power devices.

\item \textbf{SugarcaneAI: A Smartphone Application for Field Deployment} \\
We integrated the trained ShuffleNet model into a Progressive Web Application (PWA), allowing farmers to take leaf photos and get instant, explainable disease diagnoses. Grad-CAM visualizations show affected areas, such as yellow bands in banded chlorosis disease, which helps users to understand the results and take quick action. We also added an LLM-based guidance tool that offers real-time advice on diseases and suggests preventive steps.

\end{enumerate}

\autoref{fig:overall} shows an overview of our work. By providing a publicly available, detailed dataset, a fine-tuned lightweight model, and a Progressive Web Application for on-device use, our approach makes sugarcane leaf disease detection accessible in real farming conditions. This enables smallholder farmers to receive rapid, affordable, and interpretable disease diagnoses and treatment plans directly in the field.

\section{Related Works} 
Deep learning has been widely adopted in agriculture for tasks such as phenotyping, severity estimation, and disease detection across various crops. WheatSpikeNet \cite{batin2023wheatspikenet} by Batin et al. addresses wheat spike phenotyping through instance segmentation and spike counting from field images. Deeprice \cite{ritharson2024deeprice}, proposed by Ritharson et al., performs severity-based classification of rice leaf diseases. BananaSqueezeNet by Bhuiyan et al. diagnoses three prominent banana leaf diseases \cite{bhuiyan2023bananasqueezenet, arman2023bananalsd}. Other works focus on detecting diseases in citrus \cite{luaibi2021detection}, mango \cite{hossain2024deep}, tomato \cite{zhang2018can}, apple \cite{jiang2019real}, strawberry \cite{shin2021deep}, and cassava \cite{ayu2021deep}. 
\par 
Several large-scale studies have also focused on multi-crop leaf disease classification using deep learning techniques \cite{atila2021plant, mohanty2016using, yao2024deep, islam2025plantcarenet}. In recent years, several extensive datasets have been introduced for classifying leaf diseases. The PlantVillage \cite{hughes2015open} database is widely used to develop deep learning models for classifying plant diseases and testing how well these models work. It includes over 50,000 images from about 14 crop species. The dataset mainly focuses on the leaves of fruit crops like apple, grape, and strawberry. While it also has images of crops like potato, tomato, and maize, it does not include any images related to sugarcane diseases. The PlantDoc dataset \cite{singh2020plantdoc} contains 2,598 real-field images of 13 plant species, but it also does not include sugarcane leaf diseases.

\par 
Thite et al. \cite{thite2024sugarcane} proposed the Sugarcane Leaf Dataset which consists of 6,748 high-resolution images collected from Maharashtra, India which covers 9 types of diseases along with healthy and dried leaves. Daphal and Koli \cite{daphal2022sugarcane} compiled a dataset that includes 2,569 images focusing on five categories: healthy, mosaic, red rot, rust, and yellow disease, also collected in India. Hu et al. \cite{Hu2024Customized} used two datasets from Kaggle for sugarcane leaf diseases classification. Sun and Zhou \cite{sun2023se} proposed a self-collected dataset of five sugarcane leaf diseases specifically observed in the growth period of sugarcane leaves and did not open-source the dataset due to several limitations. These datasets may not capture environmental variations like soil composition and humidity unique to regions like Bangladesh, where disease phenotypes can differ significantly. 

Convolutional Neural Networks (CNNs) have shown great promise in classifying sugarcane leaf diseases. Militante et al. \cite{militante2019sugarcane} focused on binary classification between healthy and diseased sugarcane leaves and attained 95\% accuracy using a deep learning model trained on a dataset of 13,842 sugarcane images. Daphal and Koli \cite{daphal2023enhancing} used transfer learning along with ensemble method on a self-created dataset of 2569 images and reported 86.53\% accuracy using MobileNET-V2. Synthetic data generation through Generative Adversial Network (GANs) \cite{goodfellow2014generative} have also been explored by the study SugarcaneGAN \cite{li2024sugarcanegan} paper to address data scarcity. But those works struggle with dynamic environment generally found in physical field environment and fine grained details of leaf diseases which potentially lead to model overfitting to artificial features. 

To address key gaps in the literature, including the lack of public sugarcane datasets, lightweight models, and deployable tools, we introduce three contributions. As shown in \autoref{fig:overall}, our work includes the SugarcaneLD-BD dataset, the lightweight SugarcaneShuffleNet model, and the SugarcaneAI application for on-device diagnosis and treatment planning. Together, these components enable fast, accurate, and interpretable diagnosis and treatment planning for sugarcane leaf diseases in real-world farming conditions.

\section{Materials and Methods}

In this section, we provide a complete methodology of our research work. First, we discuss the complete dataset collection process and how we combined our dataset with two other publicly available datasets. Then we broadly explain our choice CNN architectures, experimental designs and evaluation methodology to develop a fast, lightweight, edge-deployable yet accurate model for sugarcane leaf disease classification.

\subsection{Sugarcane Leaf Diseases Dataset (SugarcaneLD-BD)}

SugarcaneLD-BD dataset \cite{arman2025sugarcaneld} contains high-resolution images collected from four agroecological zones in Bangladesh. The images are labeled by experts and systematically preprocessed to create a clean, diverse dataset of sugarcane leaf diseases common in Bangladesh and the Indian subcontinent.

\subsubsection{Field Description and Geographic Location}
\begin{figure}[!t] 
\centering 
\includegraphics[scale=0.6]{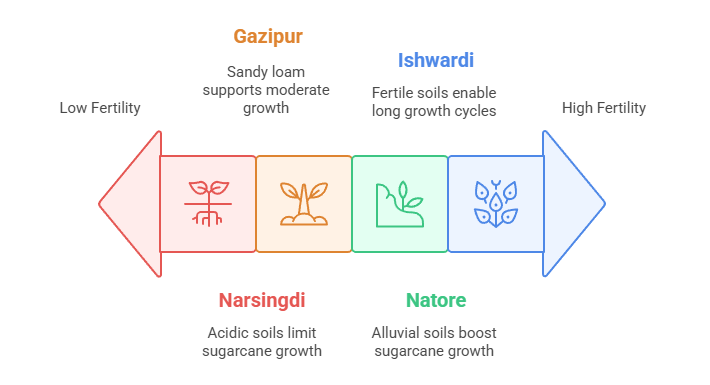} 
\caption{Sugarcane growth in four Bangladeshi regions with differing soil fertilities.} 
\label{fig:sld-growth} 
\end{figure}

This study used data from four sugarcane-producing locations in Bangladesh. These sites were chosen to reflect different agroecological conditions, farming practices, and disease prevalence. The locations are the Bangladesh Sugarcrop Research Institute (BSRI) research field in Ishwardi (Pabna), the BSRI Regional Station in Gazipur, and farmer fields in Narsingdi and Natore. \autoref{fig:sld-growth} underscores how the inclusion of four diverse agroecological zones across Bangladesh strengthens the robustness and relevance of our study. The geographic, climatic, and agronomic characteristics of these sites, along with the selection criteria, are described below.

Ishwardi, Pabna (24.130000° N, 89.060000° E) is in northwestern Bangladesh. It is near the Ganges River. The area has fertile alluvial soils. This supported sugarcane farming in the past. Ishwardi lies in the Active Ganges Floodplain (Agroecological Zone 10). The soils have high organic content and medium to high fertility. The climate is tropical wet-and-dry (Aw). It has an annual mean temperature of 28.95°C and 71.24 mm of rainfall. Moisture can stress crops. But the conditions allow longer sugarcane growth cycles.

Gazipur (23.999941° N, 90.420273° E) is in central Bangladesh. It lies in the Middle Meghna River Floodplain (Agroecological Zone 16). The soils are sandy loam and clay. They have medium fertility and moderate organic matter. The climate is typical of the country. The annual mean temperature is 28.95°C. A clear wet season supports sugarcane growth and disease spread. The BSRI Regional Station is located here. It tests disease resistance in advanced cultivars under managed conditions.

Narsingdi (23.920700° N, 90.718800° E) is in the east. Natore (24.411000° N, 89.012000° E) is in the northwest. Narsingdi lies in the Old Meghna Estuarine Floodplain (Agroecological Zone 15). Its soils are acidic with low to medium fertility. Natore is in the High Ganges River Floodplain (Agroecological Zone 11). It has nutrient-rich alluvial soils and frequent flooding. These regions share a tropical climate with Ishwardi and Gazipur. However, soil structure and moisture retention differ. This affects disease sensitivity.

\subsubsection{Image Acquisition System and Data Preprocessing}
The dataset was collected using two smartphones: Realme 8 and Xiaomi Redmi Note 11. Their specifications, including resolution, color space, focal length, aperture, exposure time, and ISO, are listed in \autoref{tab:camera_specs}. Consistent device settings and specifications ensured uniform image quality across all conditions, supporting the reliability of the dataset. Data collection took place in September and October 2023. Images were captured under varying lighting conditions, at different times of day, from multiple angles, and with diverse backgrounds to reflect field variability and support model generalization. All images were carefully labeled by qualified plant pathologists.

\begin{table}[!htpb] 
    \centering
    \small    
    \begin{tabular}{@{}lcc@{}}
        \toprule
        \textbf{Specification}       & \textbf{Camera 1} & \textbf{Camera 2} \\ \midrule
        Manufacturer              & Realme            & Xiaomi            \\
        Model               & Realme 8           & Redmi Note 11     \\
        Resolution                 & 2608 $\times$ 4624 & 2296 $\times$ 4080 \\
        Color space                & RGB               & RGB               \\
        Focal length               & 4.71 mm           & 4.25 mm           \\
        F-number                   & f/1.8             & f/1.8             \\
        Exposure time              & 1/104 s           & 1/120 s           \\
        ISO                        & 100               & 58                \\ \bottomrule
    \end{tabular}
    \caption{Specifications of the two smartphones used for datasets image aquisition}
    \label{tab:camera_specs}
\end{table}

The raw dataset contained high-resolution images with two different resolutions, as shown in \autoref{tab:camera_specs}. All images were resized to 224 × 224 pixels, a common input size for CNN architectures \cite{krizhevsky2017imagenet, simonyan2014very}. This resizing simplified the input pipeline and reduced computational complexity, making the training process more efficient \cite{he2016deep, tan2019efficientnet}.

\subsubsection{Sugarcane Leaf Diseases proposed in SugarcaneLD-BD Dataset}

\begin{figure*}[!htpb]
  \centering
  \begin{subfigure}[b]{0.24\linewidth}
    \includegraphics[width=0.8\linewidth]{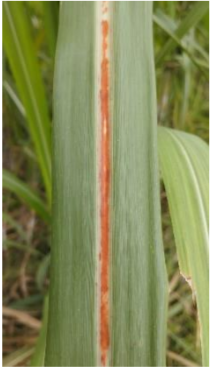}
    \caption{Red Rot}
    \label{fig:red_rot}
  \end{subfigure}
  \hfill
  \begin{subfigure}[b]{0.24\linewidth}
    \includegraphics[width=0.8\linewidth]{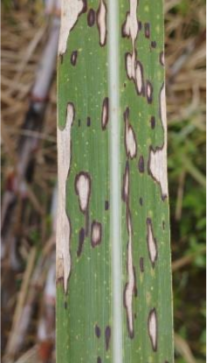}
    \caption{Ring Spot}
    \label{fig:ring_spot}
  \end{subfigure}
  \hfill
  \begin{subfigure}[b]{0.24\linewidth}
    \includegraphics[width=0.8\linewidth]{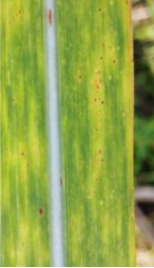}
    \caption{Red Leaf Spot}
    \label{fig:red_leaf_spot}
  \end{subfigure}
  \hfill
  \begin{subfigure}[b]{0.24\linewidth}
    \includegraphics[width=0.8\linewidth]{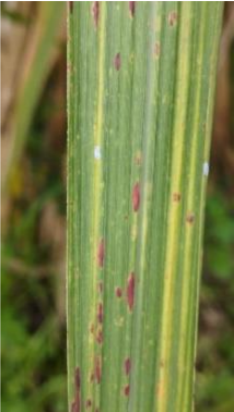}
    \caption{Eye Spot}
    \label{fig:eye_spot}
  \end{subfigure}
  \caption{Visual manifestations of sugarcane leaf diseases: (a) Red Rot, (b) Ring Spot, (c) Red Leaf Spot, and (d) Eye Spot}
  \label{fig:disease_figures}
\end{figure*}

The SugarcaneLD-BD dataset \cite{arman2025sugarcaneld} includes images of four major sugarcane leaf diseases: Red Rot, Ringspot, Red Leaf Spot, and Eye Spot. The characteristic symptoms of sugarcane leaf diseases such as Red Rot, Ring Spot, Red Leaf Spot, and Eye Spot are illustrated in Figure \ref{fig:disease_figures}.

The detailed characteristics and visual features of each disease are described below.

\noindent \textbf{1. Red Rot:} 
In Bangladesh, red rot is the most damaging sugarcane disease. It is caused by the fungus Colletotrichum falcatum. The impact depends on the climate, the strain of the pathogen, and the variety of sugarcane. In severe cases, it can reduce cane weight by up to 29\% and sugar recovery by 31\% \cite{hossain2021phylogenetic}. Red rot can look different depending on where it infects the plant. The stalk may show a straw-colored center with reddish-brown edges. In young shoots, often called 'tiller red rot', the leaves turn straw-colored and growth is stunted. If the midrib is infected, it may turn completely red. Another symptom is the appearance of red and white patches in the stalk, sometimes with alcoholic smell. The disease spreads mainly through soil and infected planting material. It can also spread through irrigation water, rain, wind, dew on lesions, and other field conditions \cite{hossain2020current}.

\noindent \textbf{2. Ring Spot:} 
Ring spot disease in sugarcane is caused by the fungal pathogen Leptosphaeria sacchari. It is a common disease found worldwide in both germplasm collections and commercial fields. Despite its wide distribution, its impact is minor because it mainly affects ageing leaves. Accurate identification is important for efficient pesticide use. The disease begins as small, oval spots that can grow into larger lesions with reddish-brown borders. These lesions often cause leaf necrosis and discoloration. Ring spot spreads through spores carried by wind or rain \cite{rott2016sugarcane}.

\noindent \textbf{3. Red Leaf Spot:} 
Red leaf spot is caused by the fungus Dimeriella sacchari. It is common in many sugarcane varieties in Bangladesh. Under favorable weather, severe outbreaks can occur. These lead to stunted plant growth and reduced sucrose content in the juice. The disease begins with small red spots (0.5–1.0 mm) on leaves. As it spreads, the spots merge into larger lesions across the leaf surface. The leaf tip is especially vulnerable. In sunlight, lesions may develop rhizoidal margins.

\noindent \textbf{4. Eye Spot Disease:} 
Eye spot is caused by the fungus Bipolaris sacchari (formerly Helminthosporium sacchari). It is a common disease in sugarcane. Although widespread, it is generally considered minor, as most varieties show natural resistance without the need for special screening. However, in susceptible cultivars, it can cause significant economic loss. The disease is marked by long lesions on leaves with red centers surrounded by a narrow ring of yellow (chlorotic) tissue. Reddish-brown streaks may also appear at the infection site and spread toward the leaf tip.

\subsubsection{Dataset Composition and Structure}
\begin{figure}[!htpb]  
    \centering
    \includegraphics[scale=0.4]{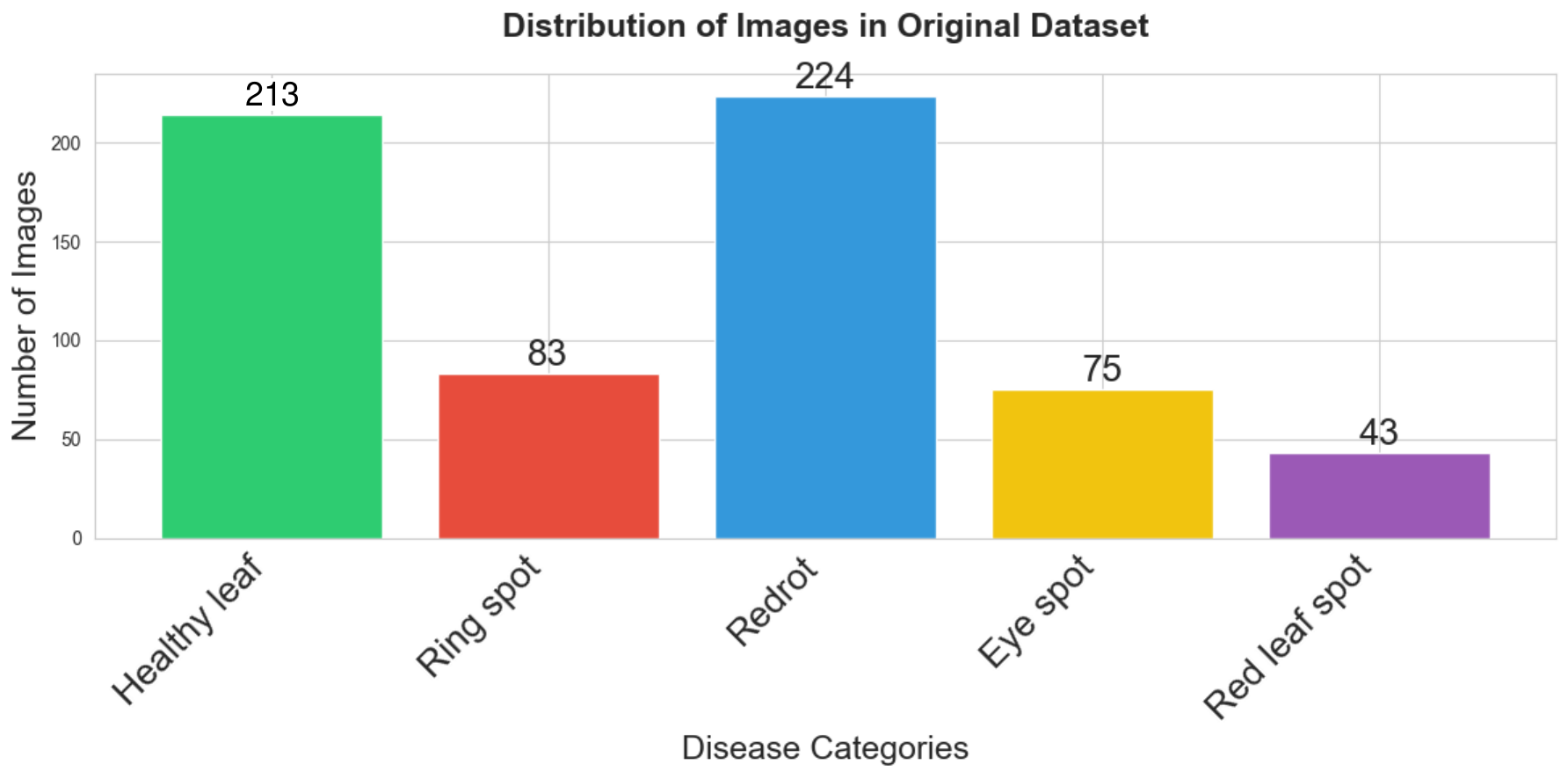}
    \caption{SugarcaneLD-BD Dataset Class Distribution}
    \label{fig:sld_bd-dist}
\end{figure}

To address the high prevalence of red rot, eye spot, and ring spot, which are serious sugarcane leaf diseases in Bangladesh \cite{hossain2021phylogenetic}, we developed the SugarcaneLD-BD dataset \cite{arman2025sugarcaneld}. It includes 638 images from four major local disease classes, all carefully labeled by plant pathologists for reliability (see \autoref{fig:sld_bd-dist}).

\subsection{Combined Sugarcane Leaf Diseases Dataset}

Many South Asian plantations, including those in Bangladesh, face diseases like yellow leaf, smut, and brown rust. We merged SugarcaneLD-BD dataset \cite{arman2025sugarcaneld} with two public datasets from India. Table~\ref{tab:data_sources} shows the dataset sources, locations, disease classes, and image counts.

\begin{table}[!htpb]
\centering
\small
\caption{Data Sources}
\label{tab:data_sources}
\begin{adjustbox}{width=\columnwidth}
\begin{tabular}{c c c c c c c}
\toprule
\textbf{Dataset} & \textbf{Location} &\textbf{Number of Classes} & \textbf{Number of Images} \\
\midrule
Thite et al. \cite{thite2024sugarcane}          & India        & 11 & 6748 \\
Daphal and Koli \cite{daphal2022sugarcane}       & India        & 5  & 2521 \\
\textbf{SugarcaneLD-BD (Ours)}                   & Bangladesh   & 5  & 638 \\
\bottomrule
\end{tabular} 
\end{adjustbox}
\end{table}

The dataset by Thite et al. \cite{thite2024sugarcane} contains 11 classes. Nine classes cover sugarcane leaf diseases, including mosaic, chlorosis, smut, brown spot, pokkah boeng, brown rust, yellow leaf, sett rot, and grassy shoot; the other two include healthy and dried leaves. The dataset by Daphal and Koli \cite{daphal2022sugarcane} includes five classes, with four disease classes: mosaic, red rot, rust, and yellow leaf. Our SugarcaneLD-BD dataset has five classes, including four diseases: ring spot, red rot, eye spot, and red leaf spot, and one healthy class.

\begin{figure*}[!htpb]
    \centering
    \includegraphics[width=0.85\textwidth]{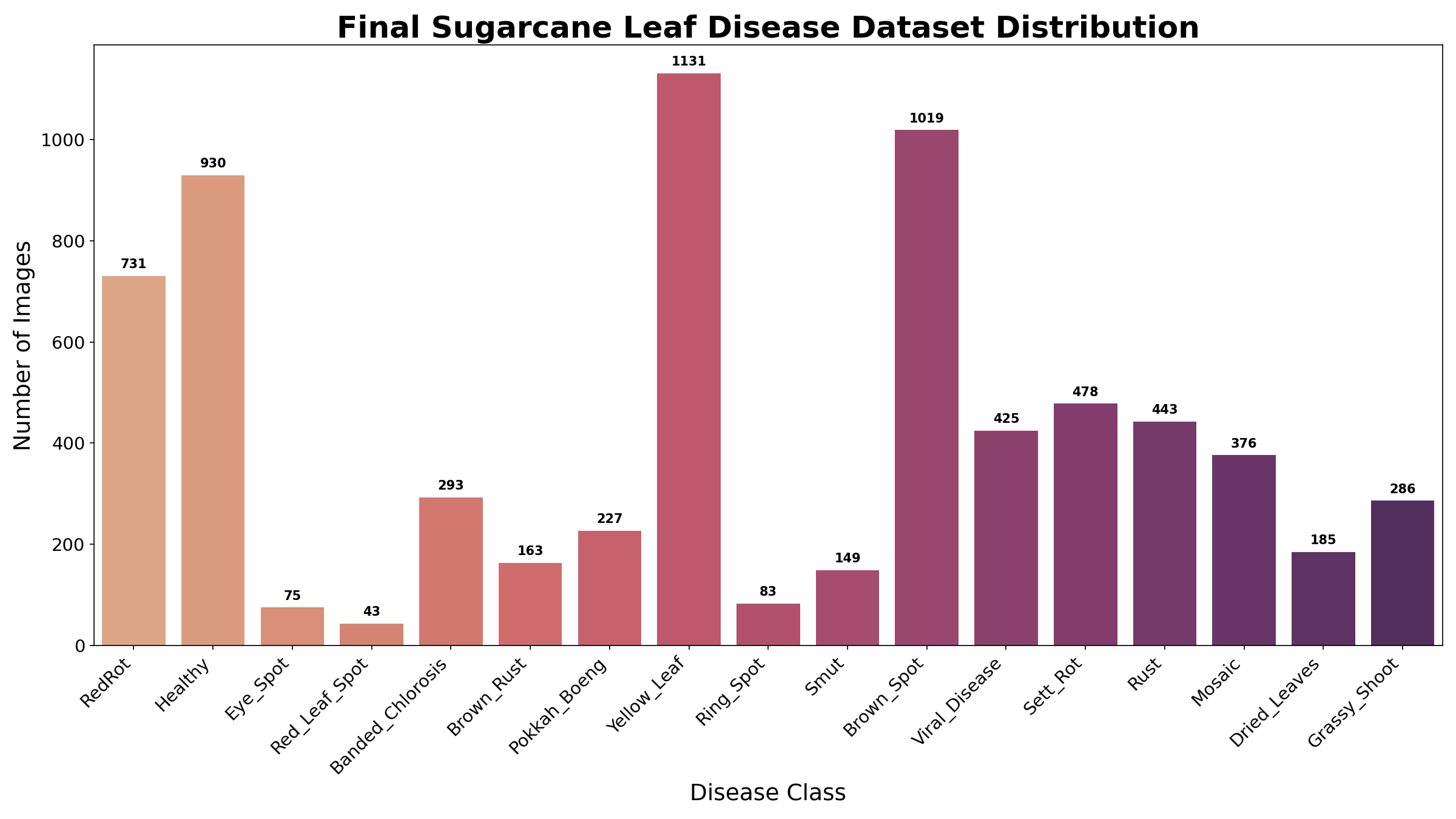}
    \caption[scale=1]{Class distribution of the combined dataset}
    \label{fig:combined-dist}
\end{figure*}

The combined dataset comprises 17 distinct classes with a total of 9,908 images. Figure \ref{fig:combined-dist} illustrates the distribution of disease classes in the dataset. The dataset is highly imbalanced, with some classes like Brown Spot and Yellow Leaf being over-represented with 1,722 samples and 1,699 samples respectively.
In contrast, classes like Red leaf spot have 43 samples, Eye spot has 75 samples, and Ring Spot has 83 samples, which are very few comparatively. This imbalance may bias model performance and highlights the importance of applying class-balancing techniques in subsequent analysis.

\subsubsection{Data preprocessing and Augmentation}

To prepare our composed sugarcane leaf disease dataset, we implemented a comprehensive preprocessing and augmentation pipeline. This process helped us address data quality issues and make the dataset suitable for robust training.
The initial dataset contained duplicates and near duplicates and inconsistent naming. These issues could have biased the model due to over-representation of samples and data leakage.
So we employed MD5 hashing to detect and remove exact duplicates. First, in a dry run, we detected all duplicates and then deleted them. We also checked for near duplicates with a 5-bit difference threshold and then deleted them. With this approach, we reduced the dataset from 9,908 to 7,037 images. We eliminated 1,264 exact duplicates and 1,607 near duplicates, which ensures each image contributes unique information to the model. We renamed all images using a consistent format like ClassName\_0001.jpg to standardize the dataset.

The dataset had significant class imbalance. Red leaf spot had only 43 images, whereas Yellow Leaf had 1,131 images. To address this, we developed a smart augmentation strategy to balance the classes while preserving visual features. We calculated an augmentation factor based on class size to reduce imbalance, as shown in \autoref{tab:augmentation_strategy}. Classes with fewer than 100 images were augmented to 600 images, whereas larger classes were capped at 1200 images or 1.5 times their original count. This approach improved the class imbalance ratio from 26:1 to 3.8:1.
The dataset was split into a training set comprising 80\% of the dataset and a test set comprising 20\% of the dataset. We used stratified sampling to maintain class proportions.
Augmentation was applied only to the training set, and an additional 5691 images were generated. The test set of 1414 images was held out for testing all models, preventing data leakage, and ensuring fair evaluation.

\begin{table*}[!htpb]
\centering
\small
\caption{Augmentation Strategy for Each Class in the composed Dataset}
\label{tab:augmentation_strategy}
\adjustbox{width=\textwidth,center}
{
\begin{tabular}{lcccc}
\toprule
\textbf{Class} & \textbf{No. of Original} & \textbf{No of images on} & \textbf{No of Augmentations} & \textbf{Final Training} \\
               & \textbf{Images}           & \textbf{Train Set}       & \textbf{applied}              & \textbf{Image Count} \\
\midrule
Eye Spot & 75 & 60 & 6 & 420 \\
Red Leaf Spot & 43 & 34 & 6 & 238 \\
Ring Spot & 83 & 66 & 6 & 462 \\
Brown Rust & 163 & 130 & 4 & 650 \\
Dried Leaves & 185 & 148 & 4 & 740 \\
Smut & 149 & 119 & 4 & 595 \\
Banded Chlorosis & 293 & 234 & 2 & 702 \\
Grassy Shoot & 286 & 228 & 2 & 684 \\
Mosaic & 376 & 300 & 2 & 900 \\
Pokkah Boeng & 227 & 181 & 3 & 724 \\
Rust & 443 & 354 & 1 & 708 \\
Sett Rot & 478 & 382 & 1 & 764 \\
Viral Disease & 425 & 340 & 1 & 680 \\
Brown Spot & 1019 & 815 & 0 & 815 \\
Healthy & 930 & 744 & 0 & 744 \\
RedRot & 731 & 584 & 0 & 584 \\
Yellow Leaf & 1131 & 904 & 0 & 904 \\
\midrule
\textbf{Total} & \textbf{7037} & \textbf{5623} & \textbf{--} & \textbf{11313} \\
\bottomrule
\end{tabular}
}
\end{table*}

\subsection{Model Training and Optimization}

In our study, we focused on developing deep learning models for sugarcane disease detection that perform reliably in field conditions. This section begins by reviewing suitable CNN architectures, followed by an explanation of how Bayesian Optimization was applied in our experiments. Finally, we discuss evaluation metrics, emphasizing those appropriate for handling class imbalance in our dataset.

\subsubsection{Convolutional Neural Networks}\label{sec:cnn}
Field-deployable sugarcane leaf disease classification requires CNNs that operate within tight memory and latency constraints while capturing key visual and textural features from high-resolution images. To satisfy this constraint, we selected six architectures: ShuffleNet, MNASNet, EdgeNeXt, EfficientNet-Lite, MobileNet and SqueezeNet, whose design philosophies converge on the same goal: maximising accuracy per parameter and per Floating Point-Operations (FLOP) through aggressive compression and efficient operators. This diverse but practical selection covers modern lightweight CNNs and keeps the experiments relevant for real-world field use.

\par 
\textbf{MnasNet} \cite{tan2019mnasnet} uses neural architecture search integrating measured mobile-CPU latency into the reward function to yield a topology that balances depth and width exactly where the hardware is fastest.
\par 
\textbf{EdgeNeXt} \cite{maaz2022edgenext} couples depth-wise convolutions with a separate depth-wise transpose-attention block thus giving the network transformer-level context without bloating its footprint.  The tiniest variant of EdgeNeXt carries only \(\sim\!1.3\,\text{M}\) parameters yet still posts roughly 71\% top-1 accuracy on ImageNet-1K.
\par 
\par\textbf{EfficientNet-Lite} \cite{tan2019efficientnet} adapts the compound-scaling principle of EfficientNet for mobile deployment by substituting Swish with ReLU6, pruning early squeeze-and-excitation blocks, and retraining under post-training quantization constraints; the resulting Lite0--Lite4 variants deliver ImageNet-level accuracy while compressing models to $\leq 5$~MB and achieving sub-5~ms inference on mid-range mobile CPUs.
\par 
\textbf{MobileNet} \cite{howard2017mobilenets} refines depthwise separable convolutions with squeeze-and-excitation blocks, hard-swish activations and a NAS-derived macro-architecture which jointly optimizes accuracy and gains empirical latency.
\par 
\textbf{SqueezeNet} \cite{iandola2016squeezenet} attains AlexNet-level performance with 50× fewer parameters by replacing wide 3 × 3 filters with “fire” modules. This modules squeeze channels via 1 × 1 convolutions before selective expansion hence slashing model size and bandwidth demands.
\par 
\textbf{ShuffleNet} \cite{zhang2018shufflenet} pursues similar ends through point-wise group convolutions coupled with a channel-shuffle permutation. This strategy preserves cross-group information flow while minimizing multiply–accumulate operations.

These architectures use smart strategies to stay small and fast without losing accuracy. They compress layers, redesign blocks, add attention, and use optimization to guide design. Together, they help build models that are light, powerful, and ready for real-world field use.

\subsubsection{Model Training Process}\label{sec:bayesopt}

In this study, we fine-tuned six lightweight Convolutional Neural Network (CNN) architectures using Bayesian Optimization to identify optimal hyperparameter settings, as detailed in Table~\ref{tab:hyperparameter_range}. This process was implemented with the Optuna framework, leveraging the Tree-structured Parzen Estimator (TPE) algorithm. The primary objective was to systematically find the combination of hyperparameters that minimizes the model's validation loss after \textbf{25} epochs of training (\emph{one Optuna trial}).

We optimized eight critical hyperparameters influencing model training and generalization, as shown in Table~\ref{tab:hyperparameter_range}. These included the learning rate, controlling the step size of weight updates, and the optimizer (Adam \cite{kingma2014adam} or AdamW \cite{loshchilov2017decoupled}), adapting model weights based on the loss function. To reduce overfitting, we applied regularization through weight decay, which penalizes large weights. Two dropout rates deactivated neurons at different stages of the classification head to reduce feature co-dependence. Label smoothing discouraged over-confident predictions, and gradient clipping prevented exploding gradients for stable learning. Each model underwent 20 optimization trials to explore the hyperparameter space.

\begin{table}[!htpb]
\centering
\caption{Hyperparameter Search Space}
\label{tab:hyperparameter_range}
\begin{tabular}{lc}
\toprule
\textbf{Hyperparameter} & \textbf{Range / Options} \\
\midrule
Learning Rate (\texttt{lr}) & 1e-5 to 1e-2 (log scale) \\
Optimizer & Adam, AdamW \\
Weight Decay & 1e-6 to 1e-2 (log scale) \\
Dropout Rate 1 & 0.1 to 0.6 \\
Dropout Rate 2 & 0.1 to 0.6 \\
Freeze Ratio & 0.0 to 0.8 \\
Label Smoothing & 0.0 to 0.2 \\
Gradient Clipping & 0.5 to 2.0 \\
\bottomrule
\end{tabular}
\end{table}

The training configuration is summarized in Table~\ref{tab:training_config}. We utilized \texttt{CrossEntropyLoss} as the loss function with a mini-batch size of 32. We used a patience of 10 epochs for early stopping. Transfer learning was employed to leverage pre-trained weights, and a \texttt{CosineAnnealingLR} scheduler adjusted the learning rate dynamically. Full training ran for 100 epochs with early stopping to prevent overfitting.

\begin{table}[!htpb]
\centering
\caption{Training Configuration}
\label{tab:training_config}
\begin{tabular}{lc}
\toprule
\textbf{Training Configuration} & \textbf{Value} \\
\midrule
Loss Function & CrossEntropyLoss \\
Transfer Learning & Yes \\
Mini-Batch Size & 32 \\
Learning Rate Scheduler & CosineAnnealingLR \\
Number of Optuna Trials & 20 \\
Epochs per Trial (Optuna) & 25 \\
Full Training Epochs & 100 \\
Early Stopping & Yes \\
Early Stopping Patience & 10 \\
\bottomrule
\end{tabular}
\end{table}

\subsubsection{Model Evaluation}\label{sec:evaluation_metrics}

We used several evaluation metrics to assess our models. Model predictions are categorized as true positives (TP), true negatives (TN), false positives (FP), and false negatives (FN), which form the basis of these metrics.

Accuracy reports the overall fraction of correct predictions:

\begin{equation}
\mathrm{Accuracy}=\frac{TP+TN}{TP+TN+FP+FN}.
\end{equation}
However, with four majority classes covering nearly half of the dataset, high accuracy may still ignore rare diseases. We therefore interpret accuracy together with class-sensitive metrics to avoid misleading results.

Precision reflects how many predicted cases are correct. Low precision may lead to unnecessary treatments and wasted resources.

\begin{equation}
\mathrm{Precision}=\frac{TP}{TP+FP}.
\end{equation}
.
Recall shows how many actual cases are detected. Low recall risks missing infections, which can allow diseases like Red Leaf Spot to spread.

\begin{equation}
\mathrm{Recall}=\frac{TP}{TP+FN}.
\end{equation}

The F\textsubscript{1} score combines precision and recall to balance correct detection and missed cases. We use macro-averaged F\textsubscript{1} to give equal importance to rare and common diseases when ranking models.

\begin{equation}
F_{1}=2\,\frac{\mathrm{Precision}\times\mathrm{Recall}}
                   {\mathrm{Precision}+\mathrm{Recall}},
\end{equation}

\section{Experimental Results}

We used the unified 17-class sugarcane leaf diseases dataset for all experiments, splitting it 80\% for training and 20\% for testing.  Each backbone model hyperparameters were fine-tuned using  Bayesian search optimization. To account for class imbalance, we reported both overall metrics and per-class metrics, ensuring a fair assessment across all classes. Inference latency and model size were measured to reflect practical deployment constraints.  The rest of this section is organized into five parts: \autoref{sec:hyperparameter_optimization} presents the best-performing hyperparameters for each model; \autoref{sec:classification_performance} reports classification results; \autoref{sec:cmputationa_efficiency} evaluates computational efficiency; \autoref{sec:interpretability} provides Grad-CAM visualizations; and \autoref{sec:comparison} compares our system with recent work.

\subsection{Hyperparameter Optimization}\label{sec:hyperparameter_optimization}

The best-performing configuration for each architecture is summarized in \autoref{tab:model_hyperparameters}. These results show clear optimization patterns across models. For example, EfficientNet-Lite and EdgeNeXt achieved the lowest validation losses with comparatively small learning rates such as $8.37 \times 10^{-5}$ and $1.27 \times 10^{-4}$, respectively, which indicates their sensitivity to finer gradient updates. The AdamW optimizer was the most consistently effective choice across the models, particularly when paired with moderate weight-decay values. SqueezeNet performed the poorest, suggesting that aggressive compression alone may not be optimal for this task. Notably, ShuffleNet responded well to the optimization, securing a robust configuration. Ultimately, MnasNet produced the lowest validation error, suggesting that its optimized configuration strikes a robust balance between regularization and learning dynamics. 

\begin{table*}[!htpb]
\centering
\caption{Model Hyperparameters}
\label{tab:model_hyperparameters}
\resizebox{\textwidth}{!}{%
\begin{tabular}{lccccccccc}
\toprule
\textbf{Model} & \textbf{Val.\ Acc.\ (\%)} & \textbf{Dropout 1} & \textbf{Dropout 2} & \textbf{Freeze Ratio} & \textbf{Grad Clip} & \textbf{Label Smooth} & \textbf{Learning Rate} & \textbf{Optimizer} & \textbf{Weight Decay} \\
\midrule
MnasNet           & 98.76 & 0.429 & 0.578 & 0.210 & 0.635 & 0.038 & 7.44e-05 & Adam  & 1.26e-05 \\
EdgeNeXt          & 98.10 & 0.414 & 0.386 & 0.148 & 0.676 & 0.132 & 1.27e-04 & AdamW & 4.06e-06 \\
EfficientNet-Lite & 98.37 & 0.158 & 0.533 & 0.016 & 0.830 & 0.107 & 8.37e-05 & AdamW & 4.80e-06 \\
MobileNet         & 97.97 & 0.103 & 0.183 & 0.720 & 1.682 & 0.038 & 8.44e-05 & AdamW & 6.81e-04 \\
SqueezeNet        & 96.91 & 0.363 & 0.593 & 0.131 & 1.686 & 0.141 & 8.00e-05 & AdamW & 3.88e-06 \\
ShuffleNet        & 98.01 & 0.480 & 0.492 & 0.453 & 1.702 & 0.052 & 6.17e-04 & Adam  & 1.27e-04 \\
\bottomrule
\end{tabular}%
}
\end{table*}

\begin{figure*}[!htpb]
    \centering
    \includegraphics[width=0.6\textwidth]{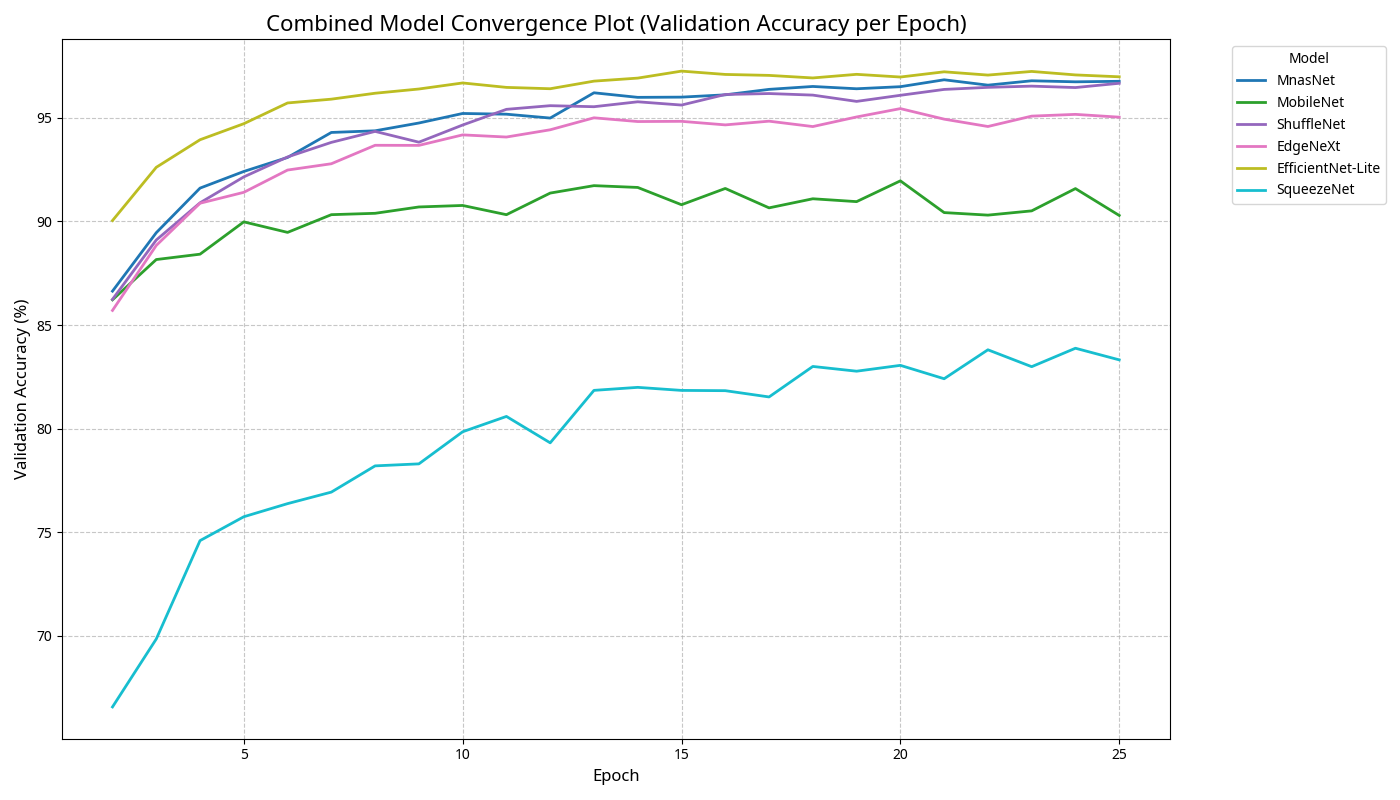}
    \caption[scale=1]{Convergence Plot}
    \label{fig:Convergence_Plot}
\end{figure*}

We further illustrate these trends in Figure~\ref{fig:Convergence_Plot}, a combined convergence plot of validation accuracy versus epoch, where MnasNet, EfficientNet-Lite, and ShuffleNet converge rapidly toward 96--97\%, while SqueezeNet lags behind---visually reinforcing the numerical findings above.

\subsection{Classification Performance}\label{sec:classification_performance}

\autoref{tab:model_performance} reports the performance of six neural network models on the combined sugarcane leaf disease dataset. Among these, ShuffleNet achieves a competitive test accuracy of 98.02\% along with strong precision (0.98), recall (0.98), and F1-score (0.98), closely rivaling the top-performing models. MnasNet records the highest test accuracy (98.51\%) and the lowest test loss (0.3360), showcasing excellent optimization. However, as will be discussed in Section~\ref{sec:cmputationa_efficiency}, MnasNet's higher computational complexity and larger model size render it less practical for edge deployment. EdgeNeXt also performs strongly with an accuracy of 98.23\%, but its elevated computational demands similarly limit its suitability for resource-constrained environments. MobileNet shows a lower accuracy (96.53\%) and test loss (0.3574), while \textbf{EfficientNet-Lite} matches ShuffleNet's accuracy (98.02\%) but incurs higher computational costs. SqueezeNet exhibits the lowest performance across all metrics, with a test accuracy of 97.31\%.

\begin{table*}[!htpb]
\centering
\small
\caption{Performance metrics of neural network models on the combined sugarcane leaf disease dataset}
\label{tab:model_performance}
\begin{tabular}{l c c c c c}
\toprule
\textbf{Model} & \textbf{Test Loss} & \textbf{Test Accuracy} (\%) & \textbf{Precision} & \textbf{Recall} & \textbf{F1-Score} \\
\midrule
\textbf{MnasNet} & 0.3360 & 98.51 & 0.98 & 0.98 & 0.98 \\
\textbf{EdgeNeXt} & 0.7693 & 98.23 & 0.97 & 0.97 & 0.97 \\
\textbf{EfficientNet-Lite} & 0.6885 & 98.02 & 0.98 & 0.98 & 0.98 \\
\textbf{MobileNet} & 0.3574 & 96.53 & 0.96 & 0.96 & 0.96 \\
\textbf{SqueezeNet} & 0.8406 & 97.31 & 0.97 & 0.97 & 0.97 \\
\textbf{ShuffleNet} & 0.3951 & 98.02 & 0.98 & 0.98 & 0.98 \\
\bottomrule
\end{tabular}
\end{table*}

\autoref{tab:per_class_f1_scores} presents the F1-scores for each disease class across the different models. ShuffleNet demonstrates robust performance across most classes, achieving F1-scores of 1.00 for several categories, such as Dried Leaves, Grassy Shoot, Pokkah Boeng, Ring Spot, and Sett Rot. It also maintains high scores (0.96-0.99) for most other classes, with a slight dip for Red Leaf Spot (0.89), a challenging class across all models. MnasNet and EfficientNet-Lite perform comparably, while MobileNet shows reduced F1-scores for classes like Mosaic (0.90) and Rust (0.93). EdgeNeXt and SqueezeNet exhibit greater variability, with EdgeNeXt struggling on Eye Spot (0.90) and SqueezeNet on Red Leaf Spot (0.88). However, all models show reduced performance on underrepresented diseases such as Red Leaf Spot, likely due to class imbalance or visual similarity with other conditions.

\begin{table*}[!htpb]
\centering
\small 
\caption{Per-class F1-scores of neural network models on the combined sugarcane leaf disease dataset}
\label{tab:per_class_f1_scores}
\begin{tabular}{l c c c c c c}
\toprule
\textbf{Class} & \textbf{MnasNet} & \textbf{EdgeNeXt} & \textbf{EfficientNet-Lite} & \textbf{MobileNet} & \textbf{SqueezeNet} & \textbf{ShuffleNet} \\
\midrule
Banded Chlorosis & 0.97 & 0.97 & 0.97 & 0.97 & 0.98 & 0.96 \\
Brown Rust & 0.97 & 0.97 & 0.97 & 0.93 & 0.97 & 0.96 \\
Brown Spot & 0.98 & 0.97 & 0.98 & 0.98 & 0.98 & 0.98 \\
Dried Leaves & 0.99 & 1.00 & 0.99 & 1.00 & 1.00 & 1.00 \\
Eye Spot & 0.97 & 0.90 & 0.97 & 0.97 & 0.94 & 0.97 \\
Grassy Shoot & 1.00 & 0.99 & 1.00 & 1.00 & 1.00 & 1.00 \\
Healthy & 0.99 & 0.99 & 0.99 & 0.97 & 0.98 & 0.99 \\
Mosaic & 0.97 & 0.97 & 0.95 & 0.90 & 0.91 & 0.97 \\
Pokkah Boeng & 1.00 & 1.00 & 1.00 & 0.99 & 0.98 & 1.00 \\
Red Rot & 1.00 & 0.99 & 0.99 & 0.96 & 0.98 & 0.98 \\
Red Leaf Spot & 0.89 & 0.82 & 0.89 & 0.82 & 0.88 & 0.89 \\
Ring Spot & 1.00 & 1.00 & 1.00 & 0.97 & 1.00 & 1.00 \\
Rust & 0.99 & 1.00 & 0.99 & 0.93 & 0.96 & 0.98 \\
Sett Rot & 1.00 & 1.00 & 1.00 & 1.00 & 1.00 & 1.00 \\
Smut & 1.00 & 1.00 & 0.98 & 0.98 & 0.98 & 1.00 \\
Viral Disease & 0.98 & 0.99 & 0.98 & 0.97 & 0.98 & 0.99 \\
Yellow Leaf & 0.97 & 0.97 & 0.96 & 0.96 & 0.96 & 0.96 \\
\bottomrule
\end{tabular}
\end{table*}

\begin{figure*}[!htpb]
    \centering
    \begin{subfigure}[b]{0.49\textwidth}
        \centering
        \includegraphics[width=0.95\textwidth]{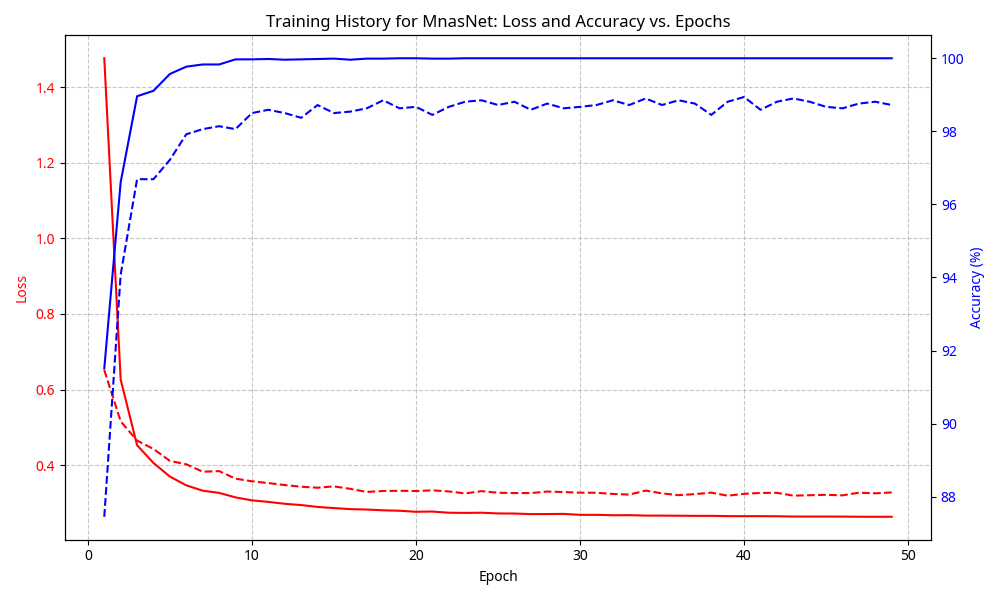}
        \caption{MNasNet}
        \label{fig:MNasNet}
    \end{subfigure}
    \hfill
    \begin{subfigure}[b]{0.49\textwidth}
        \centering
        \includegraphics[width=0.95\textwidth]{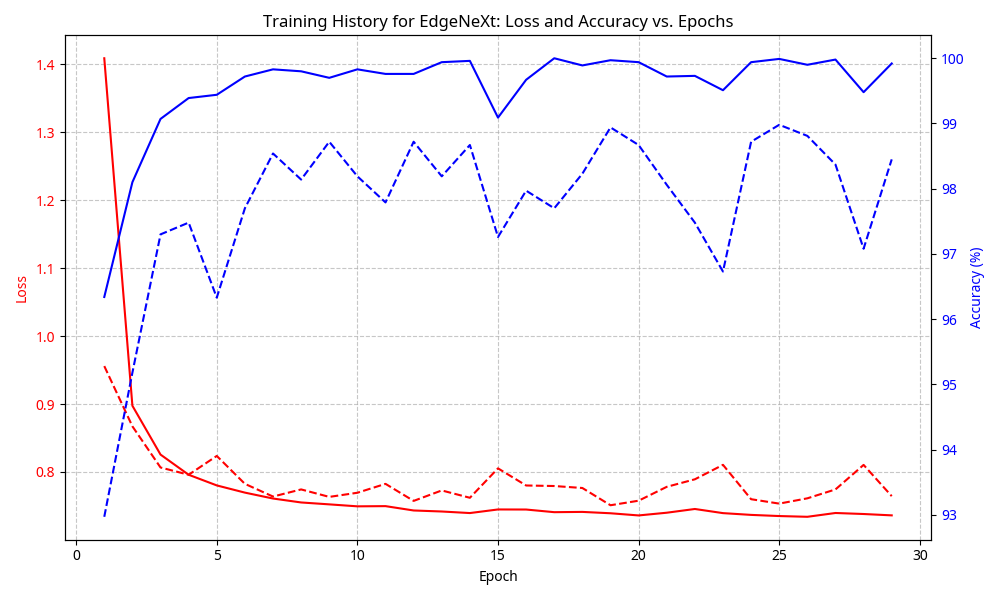}
        \caption{EdgeNeXt}
        \label{fig:EdgeNeXt}
    \end{subfigure}
    \hfill
    \begin{subfigure}[b]{0.49\textwidth}
        \centering
        \includegraphics[width=0.95\textwidth]{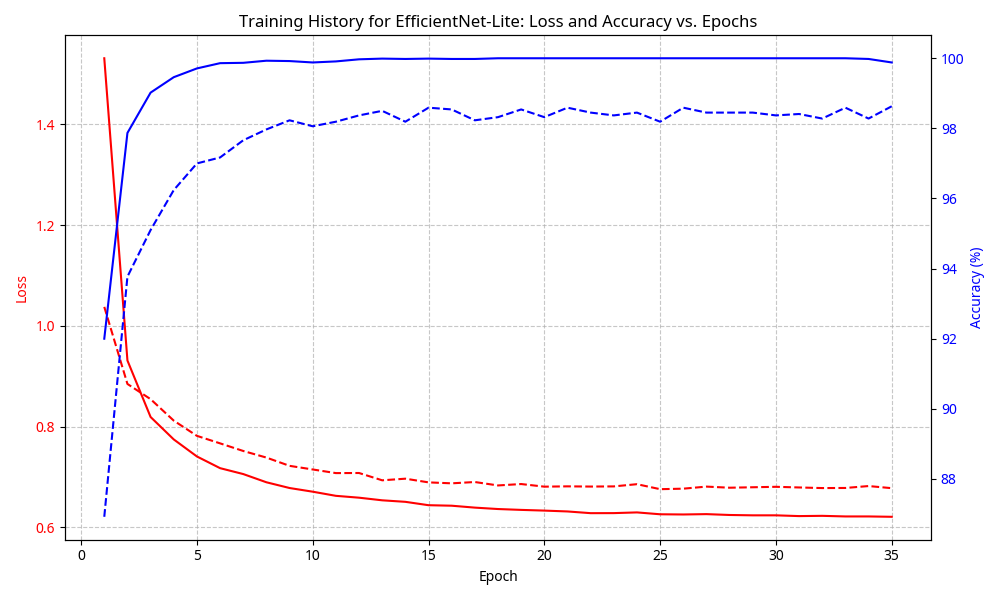}
        \caption{EfficientNet-Lite}
        \label{fig:EfficientNet-Lite}
    \end{subfigure}
    \hfill
    \begin{subfigure}[b]{0.49\textwidth}
        \centering
        \includegraphics[width=0.95\textwidth]{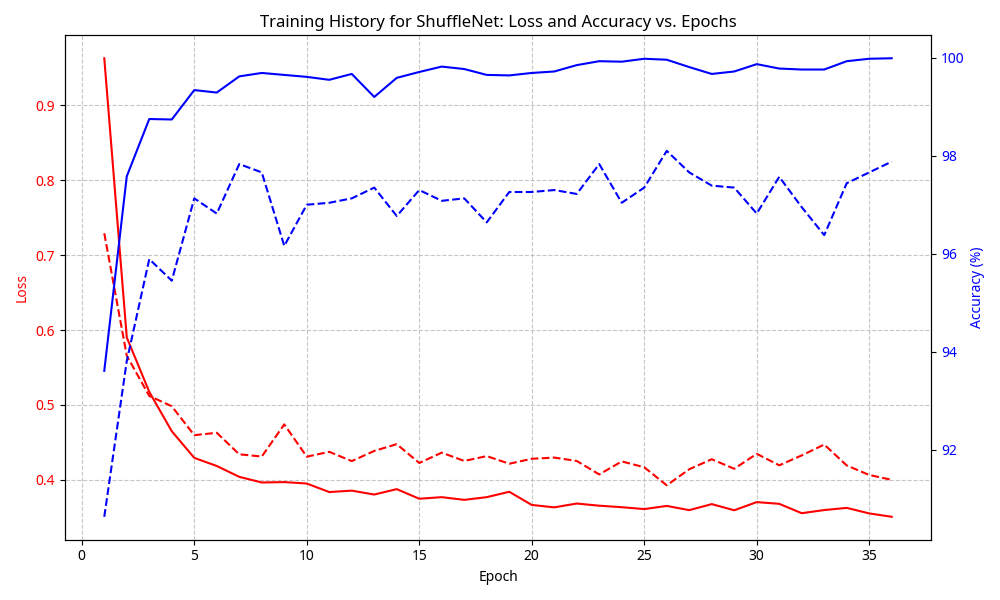}
        \caption{ShuffleNet}
        \label{fig:ShuffleNet}
    \end{subfigure}
    \caption{Training dynamics of different convolutional neural network architectures.}
    \label{fig:all_accuracies}
\end{figure*}

\autoref{fig:all_accuracies} shows the training and validation curves for four CNN models over 100 epochs. In each subplot, solid lines represent accuracy (right axis) and dashed lines represent loss (left axis). All models exhibit rapid improvement during the initial 10–15 epochs, followed by gradual changes as training progresses. ShuffleNet displays smooth and stable convergence, with consistent increases in accuracy and decreases in loss, reflecting effective optimization for this task. While overall trends are similar, some variation in curve shape and stability can be observed across models, with EdgeNeXt showing slightly more fluctuation in its loss curve.

It is evident from the results presented so far that while MNASNet and EdgeNeXt achieve marginally higher accuracies (98.51 \% and 98.23 \%, respectively), their larger model sizes and higher computational requirements, as detailed in Section~\ref{sec:classification_performance}, make them less suitable for real-time, on-device applications. In contrast, ShuffleNet offers an exceptional balance of high performance (98.02\% accuracy, 0.98 F1-score) and lightweight architecture, aligning perfectly with the goals of this research. We refer to this optimized version, tailored for sugarcane leaf-disease classification, as \textbf{SugarcaneShuffleNet}.

\begin{figure*}[!htpb]
    \centering
        \begin{subfigure}[b]{0.49\textwidth}
        \centering
        \includegraphics[width=1\textwidth]{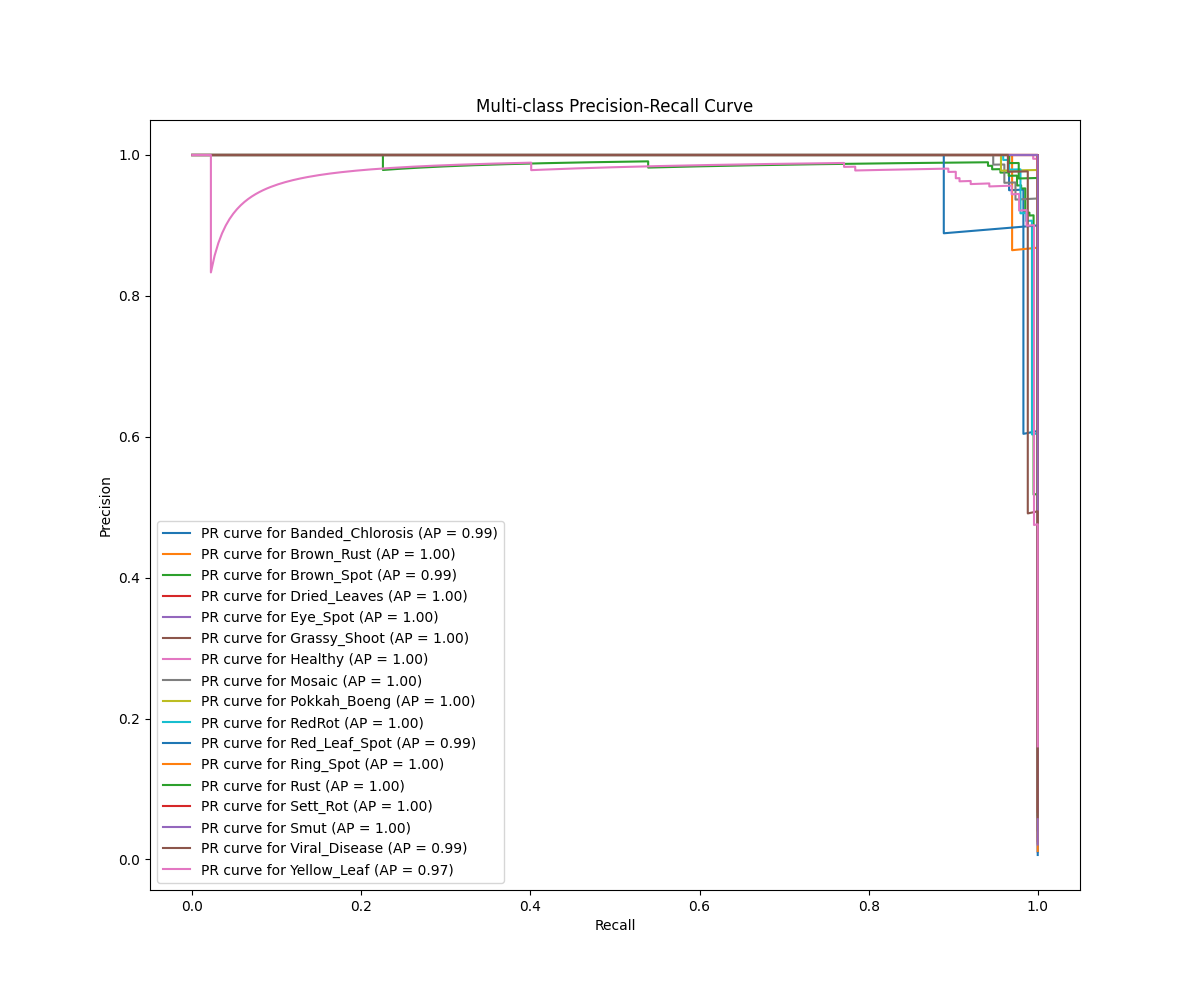}
        \caption{Precision-Recall curve}
        \label{fig:Precision-Recall curve}
    \end{subfigure}
    \hfill
        \begin{subfigure}[b]{0.49\textwidth}
        \centering
        \includegraphics[width=1\textwidth]{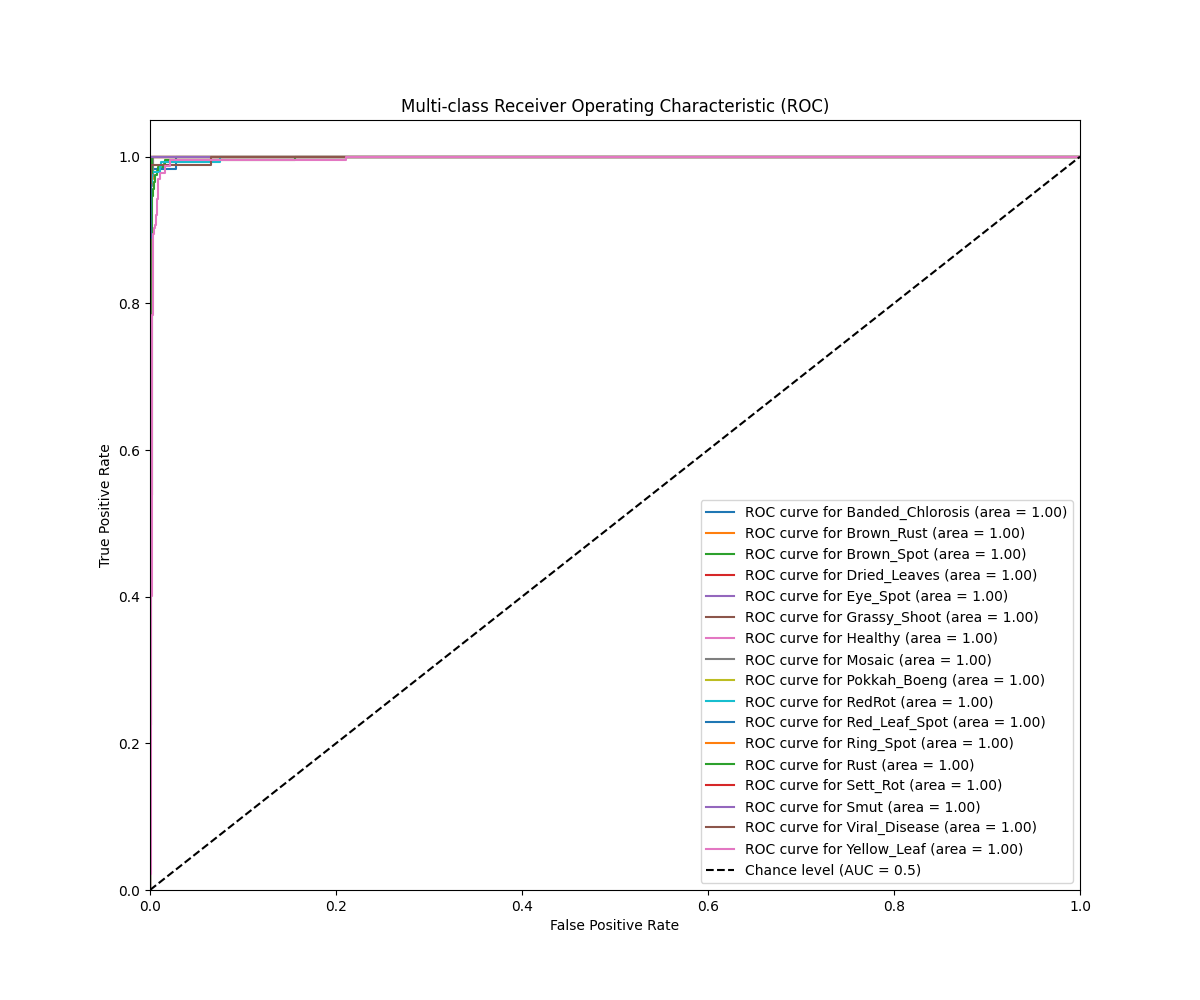}
        \caption{ROC-AUC curve}
        \label{fig:ROC-AUC curve}
    \end{subfigure}
    \hfill
    \begin{subfigure}[b]{0.95\textwidth}
        \centering
        \includegraphics[width=0.95\textwidth]{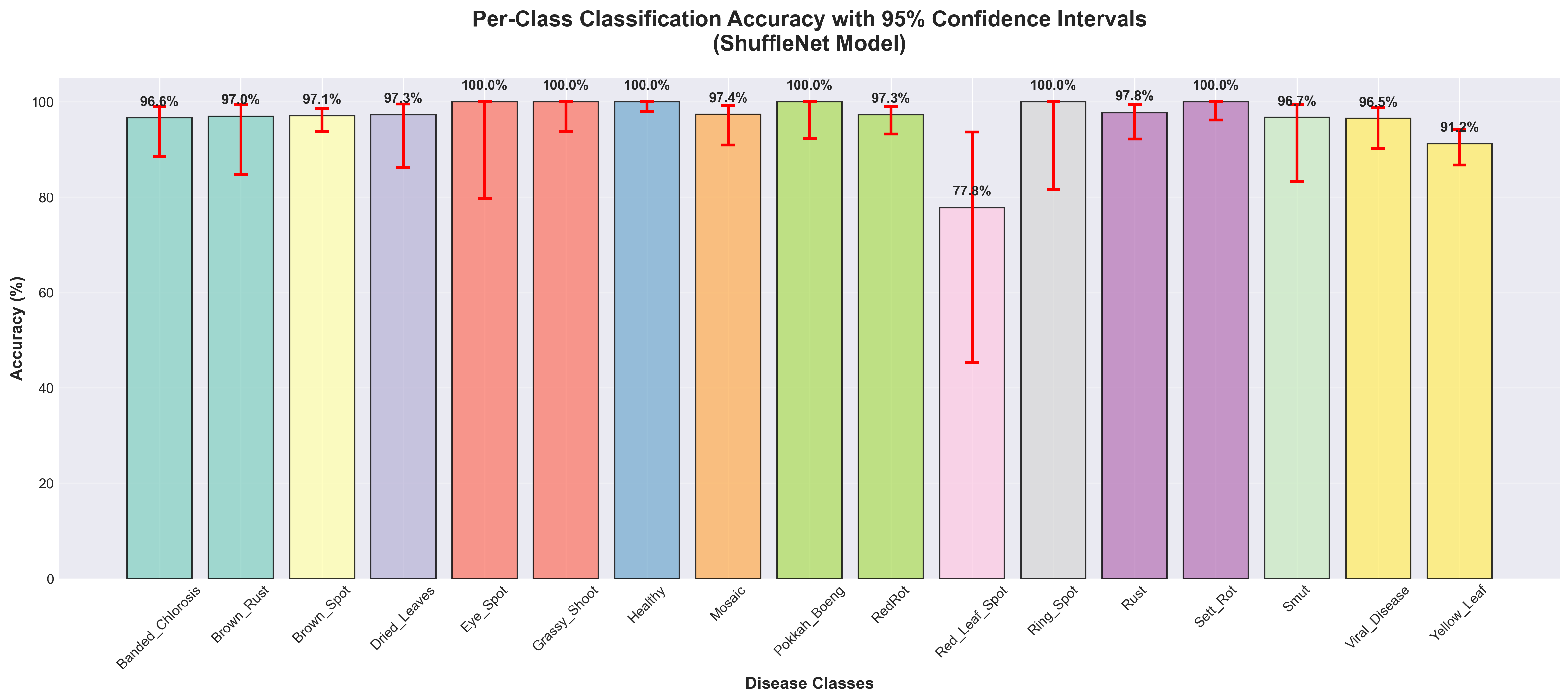}
        \caption{Class Accuracy with confidence interval}
        \label{fig:Class_Accuracy_ci}
    \end{subfigure}
    \caption{Detailed evaluation of SugarcaneShuffleNet.
(a) Precision–recall curve for multi-class classification.
(b) ROC–AUC curve for multi-class classification.
(c) Class-wise accuracy with 95\% confidence intervals.}
    \label{fig:ShuffleNet_Specific_Results}
\end{figure*}

\autoref{fig:ShuffleNet_Specific_Results} provides a detailed evaluation of SugarcaneShuffleNet across all 17 classes. \autoref{fig:Precision-Recall curve} shows the multi-class precision–recall curves. Most classes achieve high average precision, often near 1.0, reflecting strong and balanced classification performance across categories. However, a few classes, such as Red Leaf Spot and Mosaic, show sharper drops in their curves, suggesting they are harder to distinguish, possibly due to class imbalance or visual similarity. \autoref{fig:ROC-AUC curve} presents the ROC–AUC curves: nearly all classes reach AUC values close to 1.0, indicating excellent class separability. A few curves fall slightly below, but overall the model demonstrates strong discrimination across categories. \autoref{fig:Class_Accuracy_ci} displays the class-wise accuracy with 95\% confidence intervals. Most classes exceed 95\% accuracy and have tight confidence intervals, indicating consistent predictions. In contrast, classes like Red Leaf Spot and Mosaic show lower accuracy and wider intervals, indicating greater uncertainty or class imbalance. Despite these challenges, SugarcaneShuffleNet maintains reliable performance across the full set of disease categories.

\subsection{Computational Efficiency}\label{sec:cmputationa_efficiency}

As observed in \autoref{tab:neural_network_metrics}, EdgeNeXt is the most computationally expensive model, with 971.41~MMacs, a model size of 22.02~MB, and an inference time of 3.02~ms per image. In contrast, ShuffleNet (152.43~MMacs, 2.19~M parameters, 9.26~MB, 4.14~ms) and MobileNet (57.56~MMacs, 2.37~M parameters, 10.25~MB, 2.55~ms) are significantly more efficient. Although MobileNet offers the fastest inference time and the lowest MACs, its classification performance is lower than ShuffleNet's, as evidenced in \autoref{tab:model_performance}. Similarly, SqueezeNet achieves a smaller model size (7.00~MB) and fewer parameters (1.78~M), but its higher MACs (542.16~MMacs) and sub-optimal performance make it less ideal. MnasNet (324.19~MMacs, 17.59~MB) and EfficientNet-Lite (373.91~MMacs, 18.65~MB) deliver strong accuracy but are less suited for edge deployment due to their higher computational complexity and larger sizes.

\begin{table*}[!htpb]
    \centering
    \small
    \caption{Computational metrics of neural–network models on the combined sugarcane leaf disease dataset}
    \label{tab:neural_network_metrics}
    \adjustbox{width=\textwidth,center}{%
        \small
        \begin{tabular}{l c c c c}
            \toprule
            \textbf{Model} & \textbf{MACs (MMac)} &
            \textbf{Parameters (M)} & \textbf{Avg.\ inf.\ time / image (ms)} & \textbf{Model size (MB)} \\
            \midrule
            \textbf{MnasNet}             & 324.19 & 4.42 & 2.91 & 17.59 \\
            \textbf{EdgeNeXt}            & 971.41 & 5.55 & 3.02 & 22.02 \\
            \textbf{EfficientNet-Lite}   & 373.91 & 4.70 & 3.26 & 18.65 \\
            \textbf{MobileNet}           &  57.56 & 2.37 & 2.55 & 10.25 \\
            \textbf{SqueezeNet}          & 542.16 & 1.78 & 2.69 &  7.00 \\
            \textbf{ShuffleNet}          & 152.43 & 2.19 & 4.14 &  9.26 \\
            \bottomrule
        \end{tabular}%
    }
\end{table*}

We therefore select ShuffleNet for on-device deployment due to its optimal balance of computational efficiency and classification performance. Its channel-shuffle mechanism preserves information flow across grouped convolutions, avoiding the cache-unfriendly concatenations common in later lightweight architectures. With only 2.19~million parameters and an inference time of 4.14~ms per image, ShuffleNet delivers performance comparable to more resource-intensive models like MnasNet and EdgeNeXt~\cite{zhang2018shufflenet}. While its inference time exceeds MobileNet's by 1.59~ms, this difference is negligible for most edge applications, and ShuffleNet's superior classification metrics justify the trade-off. Field evaluations confirm its effectiveness for real-time disease detection on ARM Cortex-A and Raspberry Pi devices, achieving $F_1$ scores above 0.95 across multiple leaf-disease datasets~\cite{nugroho2024resource, zhou2023identification, xiang2021csms}. Additionally, its robust support in TensorFlow Lite, PyTorch Mobile, and TVM enables efficient quantisation and mixed-precision deployment~\cite{wang2025optimizing}. These attributes position ShuffleNet as an ideal backbone for handheld sugarcane disease diagnosis, ensuring fast inference while maintaining the high per-class $F_1$ scores detailed in Section~\ref{sec:classification_performance}.

\subsection{Interpretability}\label{sec:interpretability}

Transparent decision-making is crucial in crop health tools to build user trust and support informed agricultural decisions. For farmers and experts to rely on a model's diagnosis, they must understand its underlying reasoning. Explainable AI (XAI) addresses this need by revealing how the model arrives at its predictions and highlighting potential issues such as bias or spurious correlations~\cite{ferentinos2018deep, wei2022explainable, gopalan2025corn, habaragamuwa2024achieving}.

\begin{figure*}[!htpb]
    \centering
    \includegraphics[scale=0.25]{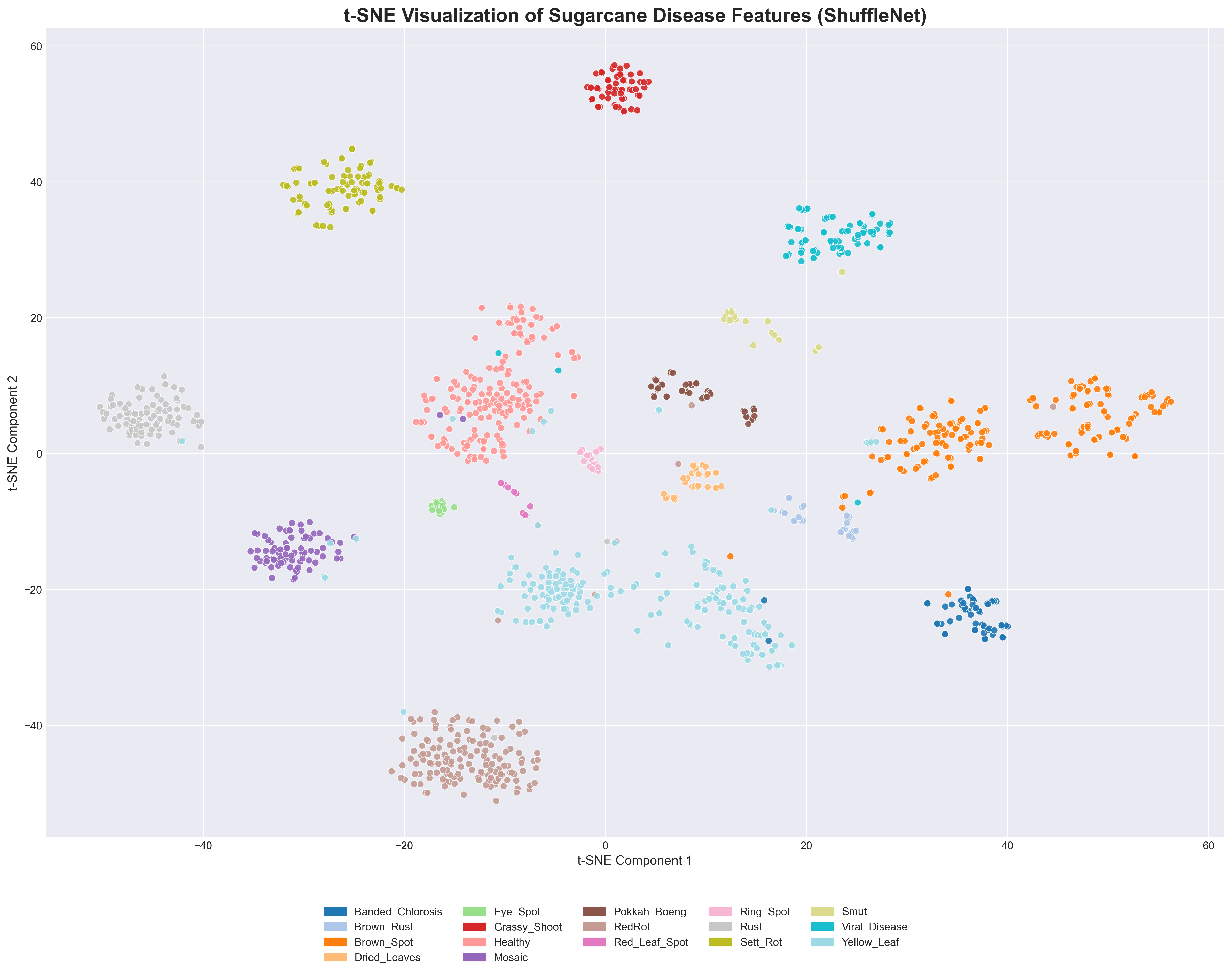}
    \caption{Visualization of class separation in feature space by SugarcaneShuffleNet.}
    \label{fig:t-sne}
\end{figure*}

We can see the model's ability to distinguish between different diseases in \autoref{fig:t-sne}. This t-SNE~\cite{maaten2008visualizing} visualization shows how the SugarcaneShuffleNet model groups similar diseases into distinct clusters. The clear separation between these clusters indicates that the model has learned to distinguish between different sugarcane diseases effectively in its feature space.

\begin{figure*}[!htpb]
    \centering
    \includegraphics[width=0.9\textwidth]{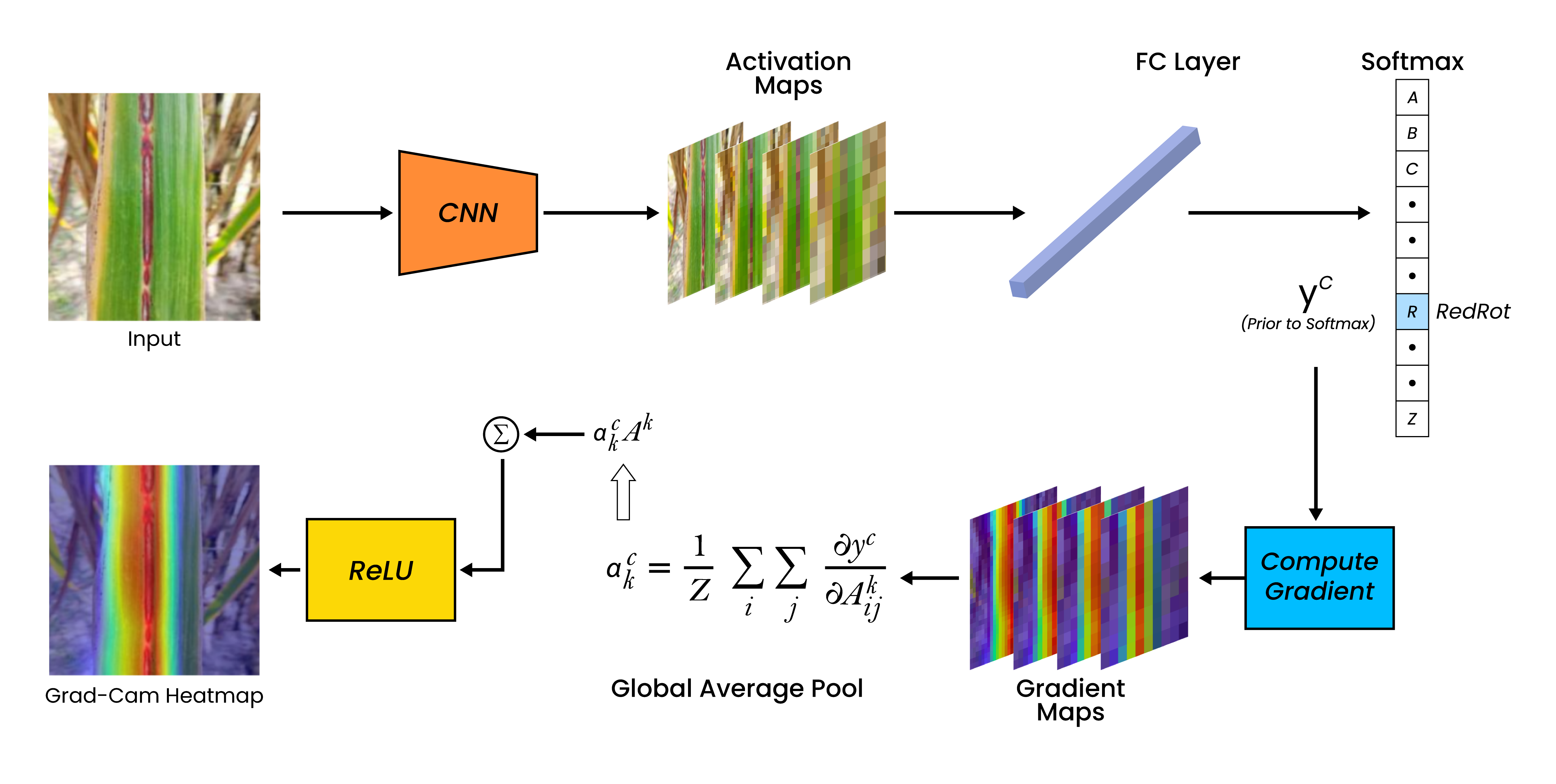}
    \caption{Grad-CAM visualization process for model prediction.}
    \label{fig:grad_cam}
\end{figure*}

To understand the reasoning behind individual predictions, we used Gradient-weighted Class Activation Mapping (Grad-CAM)~\cite{selvaraju2017grad}. It remains a standard method because of its proven ability to localize class-discriminative regions in agricultural imagery~\cite{karim2024enhancing, alhammad2025deep, nobel2024development, amara2024explainability, murindanyi2024enhanced, fahim2024comprehensive}. \autoref{fig:grad_cam} illustrates this process: a diseased leaf image is fed into the network, and Grad-CAM analyzes the model's internal features to generate a heatmap overlaid on the original image.

In this example, the Grad-CAM heatmap clearly highlights symptomatic regions on the leaf associated with red rot. We applied this method to the deployed version of SugarcaneShuffleNet. The heatmaps consistently focused on meaningful features, such as chlorotic bands, necrotic rings, or rust pustules, rather than background regions. This confirms that the model's predictions are based on actual disease symptoms and offers a clear visual explanation of its decisions.

\subsection{Comparison with Existing Works}\label{sec:comparison}

Table~\ref{tab:comparison} presents a comparative evaluation of various models on sugarcane-related classification tasks. High performance in sugarcane leaf classification is often achieved using large models or model ensembles. For example, EfficientNet-B7, fine-tuned on an 11-class dataset, achieves 99.79\% accuracy and a precision--recall score of 0.995. However, its size exceeds 250~MB and inference time is not reported, making it difficult to deploy on mobile devices \cite{srinivasan2025sugarcane}.

\begin{table*}[!htpb]
\centering
\small
\caption{Comparison with existing models}
\label{tab:comparison}
\begin{tabular}{l c c c c c}
\toprule
\textbf{Study / Model} & \textbf{\# Classes} & \textbf{Accuracy (\%)} & \textbf{Precision} & \textbf{Recall} & \textbf{F$_1$} \\
\midrule
EfficientNet-B7 \cite{srinivasan2025sugarcane} & 11 & 99.79 & 0.995 & 0.996 & 0.994 \\
SE-ViT Hybrid \cite{sun2023se} & 5 & 89.57 & 0.90 & 0.90 & 0.90 \\
Hybrid CNN + GLCM \cite{angamuthu2025comparative} & 5 & 96.80 & -- & -- & -- \\[2pt]
AMRCNN \cite{daphal2024enhanced} & 5 & 86.53 & 0.86 & 0.82 & 0.83 \\
\textbf{SugarcaneShuffleNet (Ours)} & 17 & 98.02 & 0.98 & 0.98 & 0.98 \\
\bottomrule
\end{tabular}
\end{table*}

SE-ViT combines ResNet-18, squeeze-and-excitation, and a lightweight vision transformer to achieve 89.57\% accuracy with class metrics around 0.90 on five classes, but it does not report model size or inference time \cite{sun2023se}. Angamuthu et al.\ \cite{angamuthu2025comparative} report 96.80\% accuracy on the same five classes using CNN features combined with grey-level co-occurrence matrices but do not provide precision or recall. AMRCNN is designed for mobile use with multi-level residual blocks; it achieves 86.53\% accuracy (precision 0.86, recall 0.82) on a five-class dataset, highlighting the trade-off between accuracy and latency \cite{daphal2024enhanced}.

The proposed SugarcaneShuffleNet model, despite addressing a larger 17-class problem, attains an accuracy of 98.02\%, a precision of 0.98, a recall of 0.98, and an $F_1$-score of 0.98, demonstrating competitiveness with existing approaches that generally tackle fewer classes. Most classes achieve $F_1$-scores above 0.97, with only two under-represented categories dipping below 0.90. This slight imbalance marginally affects overall metrics yet underscores the model's robustness when confronted with a broader and more challenging classification task.

\section{Progressive Web App}

We built a dual-mode system using SugarcaneShuffleNet as the classification backbone and Gemini as the recommendation engine. Through the app, farmers and field workers can receive real-time disease identification and actionable guidance.

\subsection{Real-time Inference} 

\begin{figure*}[!htpb]
    \centering
    \includegraphics[scale=0.15]{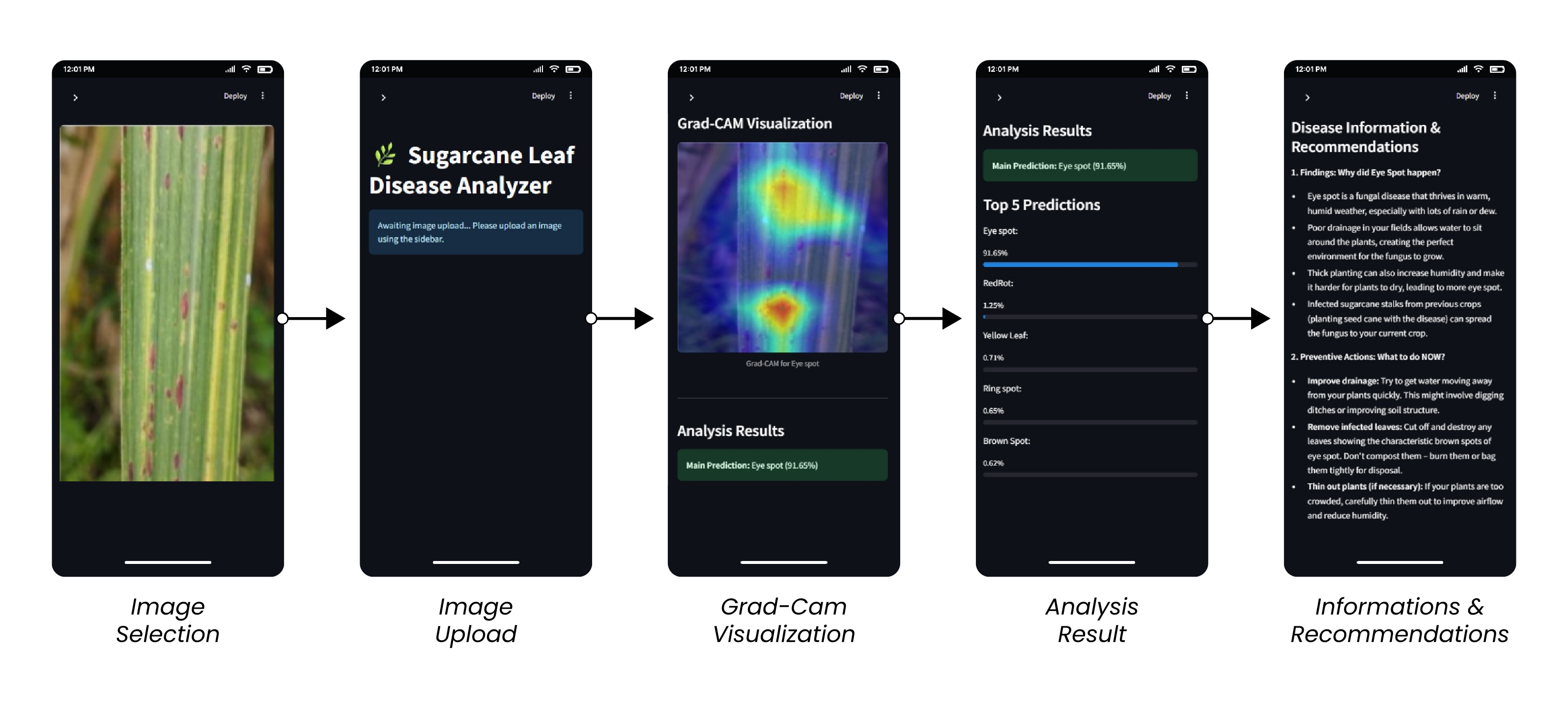}
    \caption{Workflow of the SugarcaneAI app. The app guides users through image selection, image upload, Grad-CAM visualization, disease analysis, and provides disease information and management recommendations.}
    \label{fig:sugarcane-ai-app}
\end{figure*}

Farmers and field workers can either upload an image of a sugarcane leaf or capture one in real time using the app. The deployed SugarcaneShuffleNet model processes the image and outputs class probabilities along with the top prediction and its confidence score, as shown in \autoref{fig:sugarcane-ai-app}. Simultaneously, Grad-CAM creates a heatmap that shows which parts of the leaf influenced the diagnosis, like necrotic rings for ringspot or chlorotic bands for eye-spot, helping users understand and trust the result. After showing the top prediction, the app lists the top five predicted classes with their confidence scores to help users identify possible misclassifications.

\subsection{Real-time Recommendation} 

After a disease is identified, the user taps the “Recommendations” button. This triggers a call to our Gemini API \cite{team2023gemini} with the detected disease name. The app generates a prompt asking Gemini for three sections of advice: the cause of the disease, immediate steps to limit its spread, and long-term control strategies. Gemini returns a concise text response, organized into the three requested sections. The app then displays this guidance along with the Grad-CAM heatmap within the main interface. Gemini was specifically selected for its sophisticated reasoning and synthesis capabilities. By combining visual diagnosis with instant, tailored advice, the app links symptom detection to agronomic action, addressing key gaps in field-ready systems as shown in \autoref{fig:sugarcane-ai-app}.

\section{Limitations}

Although SugarcaneShuffleNet shows promising performance on a large and diverse dataset, several limitations should be acknowledged. Despite its diversity, the dataset suffers from significant class imbalance: for instance, Brown Spot and Yellow Leaf contain 1,019 and 1,131 images, respectively, whereas Red Leaf Spot and Ring Spot have only 43 and 83 samples. This imbalance can skew learning and degrade performance on rare classes. Future work could address this by collecting more samples from under-represented categories or employing targeted augmentation strategies. Additionally, synthetic-data generation methods, such as Generative Adversarial Networks (GANs), may enhance minority-class representation and improve overall model robustness.

Current sugarcane-leaf datasets cover only a few disease stages and often originate from a single region \cite{thite2024sugarcane, daphal2022sugarcane}. Such limited coverage constrains a model’s ability to generalize. A larger dataset that spans growth stages, lighting conditions, cultivars, and geographic locations is therefore needed to improve classification reliability in heterogeneous field environments.

Finally, while our model already runs in real time with low latency, further compression is possible. Techniques such as structured pruning \cite{han2015learning}, quantisation-aware training \cite{jacob2018quantization}, and knowledge distillation \cite{hinton2015distilling} could make the network even smaller and faster, which is particularly valuable for offline or severely resource-constrained deployments.

\section{Conclusion}

This work presents an end-to-end solution for sugarcane leaf-disease classification through three main contributions. First, we developed the SugarcaneLD-BD dataset with expert-annotated images from Bangladesh and combined it with two public datasets. De-duplication and balanced augmentation yielded a diverse 17-class corpus capturing real-world variation in lighting, background, and imaging conditions. Second, we designed SugarcaneShuffleNet, a lightweight CNN for real-time diagnosis, and compared it with several other efficient architectures. Among the six tested models, the optimised SugarcaneShuffleNet achieved the best balance, reaching 98.02 \% accuracy, a macro-F\textsubscript{1} of 0.98, and fast inference with a 9.26 MB model size and 4.14 ms per image. Third, we deployed this model in SugarcaneAI, a field-ready progressive web application that provides real-time, explainable diagnoses and practical management recommendations. Together, our dataset, model, and application constitute an efficient and practical pipeline for sugarcane disease classification. Future work will focus on expanding the dataset with rarer diseases, broadening geographic diversity, and further compressing the model to enhance robustness and deployability on ultra-low-power hardware.

\section*{Data and Code Availability}

\begin{sloppypar}
The SugarcaneLD-BD dataset \cite{arman2025sugarcaneld} is publicly available on Mendeley Data at \url{https://doi.org/10.17632/n8mpzb7p4k.1}.  
A mirrored copy is also provided on Kaggle at
\url{https://www.kaggle.com/datasets/shifatearman/sugarcaneld-bd-dataset}.  
All source code covering, data-processing pipelines, model-training notebooks, inference scripts, and the SugarcaneAI progressive-web application, is hosted in a dedicated GitHub repository at
\url{https://github.com/shifatearman/SugarcaneShuffleNet}.
\end{sloppypar}

\section*{Ethics Statements}

This article does not contain any studies involving animals or human beings. No data was collected from social media platforms.

\section*{Acknowledgments}

The sugarcane disease symptoms were primarily collected at the Bangladesh Sugar Crop Research Institute regional office in Gazipur. The authors acknowledge the contribution of those who collected photographs of disease incidence. We gratefully acknowledge the support of farmers from whose fields we collected photographs of disease incidence. We would also like to acknowledge Wasique Ahmed for helping with the illustrations.

\section*{CRediT authorship contribution statement}

\textbf{Shifat E. Arman}: Conceptualization, Methodology, Data Curation, Resources, Supervision, Visualization, Writing - Original Draft. 
\textbf{Hasan Muhammad Abdullah}: Conceptualization, Data curation, Funding Acquisition, Resources, Supervision, Writing - Review \& Editing.
\textbf{Syed Nazmus Sakib}: Data curation, Investigation, Methodology, Resources, Software, Visualization, Writing - Original Draft. 
\textbf{RM Saiem}: Data curation, Resources. 
\textbf{Shamima Nasrin Asha}: Data curation, Resources. 
\textbf{Md Mehedi Hassan}: Conceptualization, Resources, Writing - Review \& Editing.
\textbf{Shahrear Bin Amin}: Resources, Writing - Review \& Editing. 
\textbf{S M Mahin Abrar}: Resources, Writing - Review \& Editing. 

\section*{Declaration of Competing Interest}

The authors declare that they have no known competing financial interests or personal relationships that could have appeared to influence the work reported in this paper.

\section*{Funding}

The authors express sincere gratitude for the financial support received through a special allocation grant from the Ministry of Science and Technology, Government of the People’s Republic of Bangladesh.

\section*{Declaration of Generative AI and AI-assisted Technologies in the Writing Process}
During the preparation of this work the authors used ChatGPT and QuillBot in order to check grammar, spelling, and to improve fluency. After using these tools/services, the authors reviewed and edited the content as needed and take full responsibility for the content of the published article.

\bibliographystyle{elsarticle-num}
\bibliography{ref}
	
\end{document}